\documentclass[10pt,twocolumn,letterpaper]{article}

\usepackage[pagenumbers]{cvpr} % To force page numbers, e.g. for an arXiv version

\usepackage{graphicx}
\usepackage{amsmath}
\usepackage{amssymb}
\usepackage{booktabs}
\usepackage[labelfont=bf]{caption}
\usepackage{subcaption}
\usepackage{multirow}
\usepackage{comment}
\usepackage{float}

\usepackage[pagebackref,breaklinks,colorlinks]{hyperref}

\usepackage[capitalize]{cleveref}
\crefname{section}{Sec.}{Secs.}
\Crefname{section}{Section}{Sections}
\Crefname{table}{Table}{Tables}
\crefname{table}{Tab.}{Tabs.}

\def\cvprPaperID{3898} % *** Enter the CVPR Paper ID here
\def\confName{CVPR}
\def\confYear{2023}

\begin{document}

\title{Learning a Pedestrian Social Behavior Dictionary}

\author{Faith Johnson\\
Rutgers University\\
{\tt\small faith.johnson@rutgers.edu}
\and
Kristin Dana\\
Rutgers University\\
{\tt\small kristin.dana@rutgers.edu}
}
\maketitle

%%%%%%%%% ABSTRACT
\begin{abstract}
   Understanding pedestrian behavior patterns is a key component to building autonomous agents that can navigate among humans.
   % such as delivery robots, robots for mobility assistance, and other autonomous agents. 
   We seek a learned dictionary of pedestrian behavior to obtain a
    semantic description of pedestrian trajectories. Supervised methods for dictionary learning are impractical since pedestrian behaviors may be unknown a priori and the process of manually generating behavior labels is prohibitively time consuming. We instead utilize a novel, unsupervised framework to create a taxonomy of pedestrian behavior observed in a specific space. 
%, such as a walkway, courtyard, or room. 
   First, we learn a trajectory latent space that enables unsupervised clustering to create an interpretable pedestrian behavior dictionary.
   %This dictionary allows key questions to be addressed, such as how pedestrians use a space. It also has utility for a simplified pedestrian trajectory prediction, conditioned on pedestrian behavior. 
   % we don't do this in the paper so lets take it out of the abstract...
  % to facilitate the localization of interesting social waypoints, i.e. significant social behavior landmarks. 
   We show the utility of this dictionary for building pedestrian behavior maps to visualize space usage patterns and for computing the distributions of behaviors. We demonstrate a simple but effective trajectory prediction by conditioning on these behavior labels. While many trajectory analysis methods rely on RNNs or transformers, we develop a lightweight, low-parameter approach and show results comparable to SOTA on the ETH and UCY datasets. 
%   With the resulting dictionary we can:
    % 1) Create pedestrian behavior maps to show how a space is used over a give point in time and an aggregate of time;
    % 2) Create a video summary from raw behavior by selecting pedestrian patterns that are interesting (“couples walking”, “people congregating”); 
    % 2) Provide trajectory prediction conditioned on behavior with very simple networks that have comparable performance to large state of the art networks;
\end{abstract}

\vspace{-25pt}
\section{Introduction}
\label{sec:intro}

\begin{figure}
    \centering
    \includegraphics[width=0.48\textwidth]{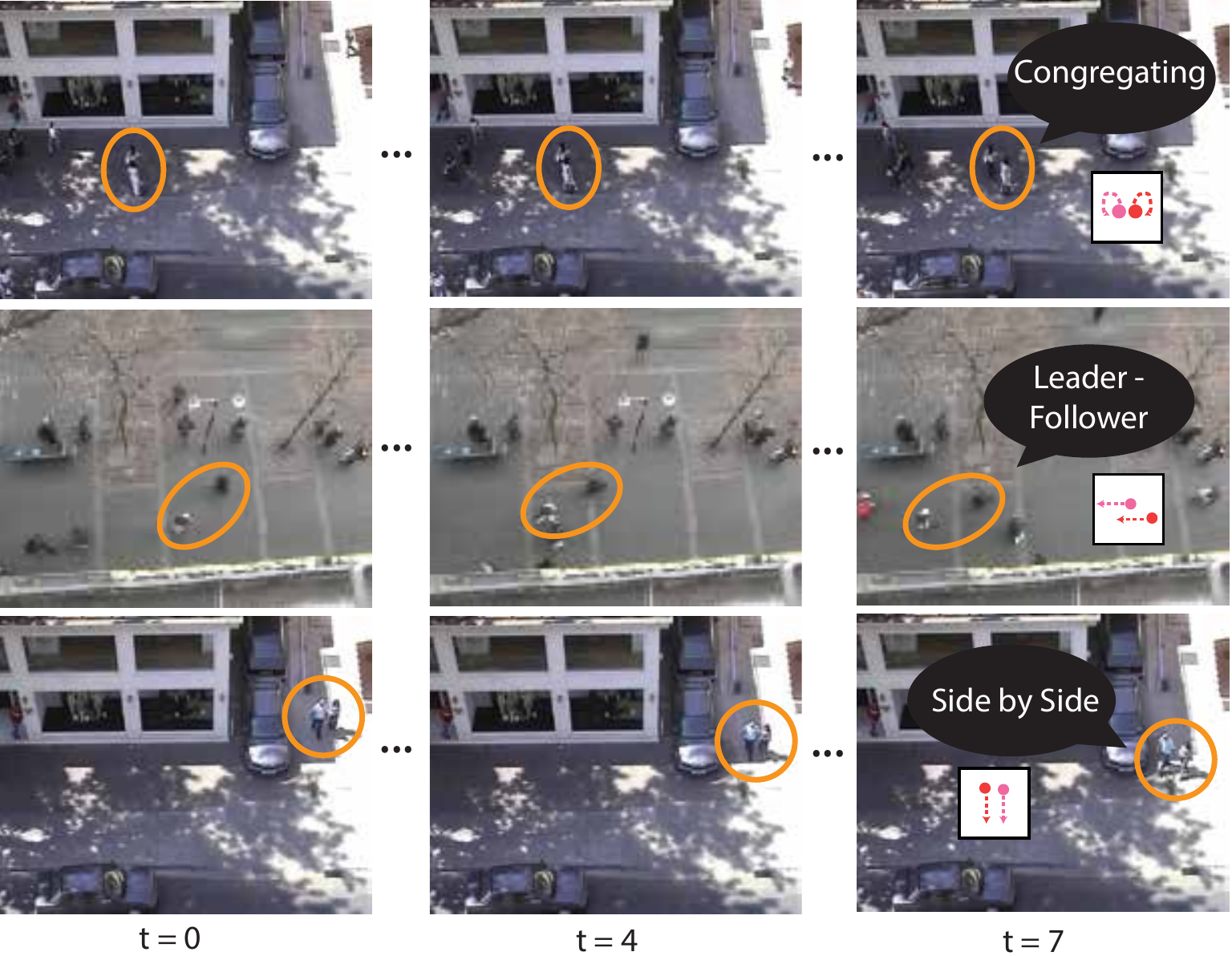}
    \caption{Predicting pedestrian trajectories becomes easier when there is an understanding of the underlying social behaviors taking place. We create PT-Net to predict these social behaviors given a set of historical pedestrian locations. Each behavior maps to a different location in an embedding space, which allows for the characterization of a spectrum of pedestrian behavior.}
    \vspace{-10pt}
    \label{fig:teaser}
\end{figure}

% - What is the problem/topic 
% good and bad of SOTA and a brief description of our method to fix it
%The successes of computer vision in robust recognition have paved the way for vision to guide autonomous agents in real world environments. 
The success of computer vision in robust recognition has paved the way for %interaction goals such as 
vision-guided autonomous agents in real world environments.
For embodied agents to navigate amongst people, an understanding of pedestrian behavior is important for maneuvering in a non-disruptive manner. 
% - What's been done before (not a full related work section just a concise overview sentence or two)
Trajectory prediction of pedestrians has received significant attention in recent years, with algorithms %have achieved great success in predicting the future steps of pedestrians by learning
that learn time-series representations of trajectories while taking into account both scene %context \cite{shafiee2021introvert} and other nearby pedestrians \cite{vemula2018social}. 
context and other nearby pedestrians.
While trajectory prediction algorithms are powerful, they %use raw trajectories as input and 
output sequences of x-y coordinates which do not provide high level, interpretable knowledge %of the scene that can be used to explain what is happening in terms of human behavior.  
of the scene or explain scene dynamics in terms of human behavior.  
%  More about the approach, emphasizing its simplicity  and unsupervised
% A key observation about 

Trajectories are already low-dimensional as a sequence of x-y values with limited long-range dependencies. Consequently, the computational tools used in past trajectory prediction such as %recurrent networks \cite{alahi2016social,xue2020poppl}, RNNs with attention, \cite{shafiee2021introvert,vemula2018social}, 
RNNs with attention \cite{shafiee2021introvert,vemula2018social} and without \cite{alahi2016social,xue2020poppl}, 
transformers \cite{yu2020spatio,giuliari2021transformer}, and  spatio-temporal graph networks \cite{yu2020spatio,kosaraju2019social} may be unnecessarily complex for the task.  %, especially if semantic behavior labels can pre-condition the predictions.
% - What are we doing that's different? 
Instead, we create \textit{PT-net}, a network that learns a \textit{pedestrian behavior dictionary} in an unsupervised, data-driven manner to provide explainable, semantic social behavior labels for pedestrian trajectories in a scene. 
These can be used to characterize global trends in space usage and pedestrian behavior habits as well as assist in downstream tasks like trajectory prediction %. Furthermore, the pedestrian behavior labels are useful in conditioning trajectory prediction 
so that very shallow %predictors 
networks can be used to predict accurate paths. 

Unsupervised methods are an integral part of our framework because %supervised methods for dictionary learning would 
they do not require prohibitively time consuming labelling of large datasets like their supervised counterparts. Moreover, the labels themselves are unknown in this domain and need to be learned from pedestrian data. We are inspired by methods \cite{zhang2017deep} that learn networks to reproduce a t-SNE-embedding using a student-teacher framework to circumvent this problem.  PT-net creates a stationary latent space embedding of trajectories that uses t-SNE clustering as its guide to create a pedestrian behavior dictionary. % by mimicking the projection into the fixed t-SNE latent space. 
This allows the dictionary to be a low-dimensional 2D latent space mapping which groups trajectories with similar social behaviors into homogeneous clusters, due to the KL divergence loss in t-SNE promoting clustering. 
The clusters in the projection space are well-defined and correspond to interpretable %trajectory 
behavior. These clusters can be readily human-labelled
with a semantic behavior
by observing a sampling of trajectories within each cluster.
%for a small number of points within each cluster. 

Unsupervised clustering to discover social behaviors avoids the shortcomings of manually defining presumed behaviors.  For example, consider the cluster that corresponds to {\it leader-follower behavior} where two pedestrians travel approximately the same path, one in front of the other, separated by a distance. This behavior cannot be easily manually defined a priori because the inter-person distance is randomly distributed and varies among environments.  Our approach clusters behaviors in an unsupervised manner without the explicit definition of these properties, supporting the discovery of the diverse behaviors within a specific environment. A network is then constructed to mimic the unsupervised clustering and embed trajectories into a latent space where new trajectory instances can be projected. 

%The human-given, semantic names to these clusters are provided in Table \ref{tab:behaviorTaxonomy} and include ``walking as a pair'', `` stopping to congregate'', ``leader-follower pairs", and other group behaviors. %\textbf{I got rid of this table potentially? but even these examples need to be updated. I'll do it once the results section is more mature to make sure that they match} ***Include this somewhere else***

From this clustering, we can also map semantic pedestrian behaviors from the dictionary to specific locations in scenes resulting in a {\it pedestrian behavior map} that can answer key questions about an environment.
How do pedestrians use this space? Where do people congregate and socialize? What are the dominant patterns in pedestrian behavior? These questions allow the localization of interesting social waypoints, i.e. points of social behavior inflection or change in the environment, which can be used in robot motion planing. 
For example, just as a robot should avoid running into a mailbox, it should also %avoid or be navigationally aware 
be cognizant of social waypoints such as patios, % on a campus, % where students may tend to congregate to socialize, 
bus-stops, % where people are waiting, 
%paths where groups tend to walk together, and 
map kiosks, % where individuals pause to view information. 
and other places where people congregate and socialize. While one approach may be to determine these waypoints %by looking at 
using what is physically present in the space, another method is to observe how people use the space enabled by the taxonomy of an interpretable pedestrian behavior dictionary. 

Our framework has the advantage of being light-weight, unsupervised, and relying on very basic networks to achieve useful characterizations and accurate predictions of pedestrian behavior.
We provide results showing the utility of our method on the ETH \cite{ETH} and UCY\cite{UCY} datasets through the creation of pedestrian behavior maps to characterize environment usage and trajectory prediction that is conditioned on the predicted social semantic behavior using PT-net. While trajectory prediction is not our main goal, we demonstrate that our pedestrian behavior dictionary  simplifies the task of pedestrian trajectory prediction 
and maintains high performance. After predicting the social behavior cluster for a particular trajectory, the dictionary label is used to constrain the future trajectory prediction. Conditioned on the behavior, trajectories can be predicted with shallow MLPs, % instead of the usual large sequence networks. In this task, we 
which show comparable performance with state of the art (SOTA) methods that use much larger, more complex networks.

% within a shallow network and selects an MLP to make the future trajectory prediction. Each cluster has one MLP that is trained to make predictions on that specific trajectory type. This effectively narrows the potential state space of the prediction problems to allow the simple networks to complete this historically complex task. Training a scene-specific model inherently accounts for scene context.  While some applications require no pre-training on the scene, the observation of a scene before algorithm deployment is quite reasonable in numerous applications, such as ..., due to a ubiquitous fixed camera. 
%I really don't want to say surveillance here, but I cannot think of anything else to say
%

% Explain the results
 
%of the following tasks:
% \textbf{1)} Create pedestrian behavior maps to show how a space is used over a give point in time and an aggregate of time;
% \textbf{2)} Create a video summary from raw behavior by selecting pedestrian patterns that are interesting (“couples walking”, “people congregating”); 
% \textbf{3)} Provide trajectory prediction conditioned on behavior with very simple networks that have comparable performance to large state of the art networks.
% \textbf{not a fan of the double lists here, but I guess it's fine for now}
% Enumerate the papers contributions
Our main contributions are threefold: \textbf{(1)} introduction of PT-net: an unsupervised  method for computing a semantically meaningful {\it pedestrian behavior dictionary} using a novel t-SNE imitator network; \textbf{(2)} construction of interpretable {\it pedestrian behavior maps} to characterize environment usage patterns in terms of pedestrian behavior; \textbf{(3)} competitive pedestrian trajectory prediction results with a much simpler network than current SOTA.
\begin{figure*}
    \centering
    \includegraphics[width=\textwidth]{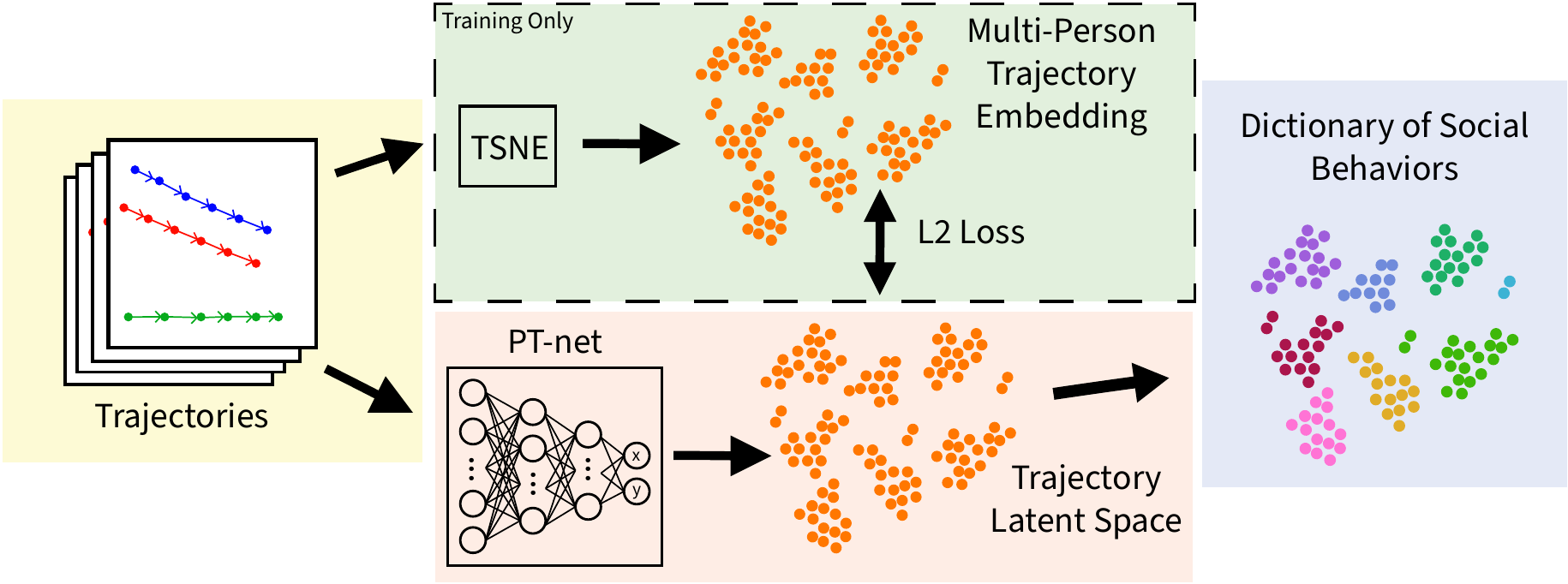}
    \caption{To make the pedestrian behavior dictionary, trajectories for small groups of nearby pedestrians are extracted from the datasets in overlapping windows of a fixed length. Velocity-based and proximity-based features are computed from these trajectories. During training, these processed trajectories are input to the t-SNE algorithm to create a 2D trajectory latent space embedding. PT-net takes in the processed trajectories and directly predicts the corresponding latent space coordinates from the ground truth t-SNE embedding. This learned coordinate embedding separates distinct pedestrian behavior into clusters in the space, each of which comprises a social behavior to form a pedestrian behavior dictionary. During inference, the processed trajectories are directly input to PT-net to get the embedding coordinates which are matched to the closest social behavior cluster.}
    \vspace{-5pt}
    \label{fig:process}
\end{figure*}

\section{Related Work}

% Prior Models use RNN, LSTM and transformers, 
In prior work, pedestrian trajectories are often modelled %as sequences 
with standard sequence based methods such as  LSTMs\cite{alahi2016social,xue2018ss,shafiee2021introvert}, and more recently transformers \cite{yu2020spatio,giuliari2021transformer}, with the primary goal of %of prior work has been
pedestrian trajectory prediction, not behavior dictionaries. 
% as their attention module and sequence modeling capabilities lend themselves well to the pedestrian trajectory prediction problem.
Other computational frameworks for trajectory prediction include GANs\cite{gupta2018social,sadeghian2019sophie,kosaraju2019social},  graph networks\cite{vemula2018social,mohamed2020social,li2020evolvegraph}, and CVAEs\cite{lee2022muse,chen2021personalized,mangalam2020not}. Some methods explicitly model the social interactions between pedestrians using attention \cite{kothari2021human,vemula2018social} or contrastive learning \cite{liu2021social} 
in order to enhance future trajectory prediction.
Unlike prior work, our emphasis is on
%not trajectory prediction, nor individual trajectories. Instead, our goal
building a short-term pedestrian behavior dictionary answering basic questions like: What behaviors do small groups of pedestrian exhibit in the space? The resulting description of interpretable pedestrian behavior can be used to characterize the usage of a space by quantifying distributions and locations of pedestrian behavior.

% We don't do sequence modeling
In our work, a dictionary of behaviors is learned from small groups of pedestrians over fixed time windows in a computationally lightweight approach without sequence models. 
%Our approach is unique compared to sequence-based trajectory models  because of its simplicity. %By utilizing a cluster-based representation, behavior-conditioned representations can be modelled concisely. 
Recent work in trajectory clustering \cite{tokmakov2020unsupervised}  shows the advantage of  clustering-based, unsupervised representations for trajectories in the domain of action recognition, using video as inputs.  Prior work in trajectory clustering \cite{Ge12visionbased, zhang2016red,han2019pedestrian} also show the utility of clustering to analyze dependencies in crowded scenes. 
Our  trajectory framework uses small groups of image coordinates as input to account for short-term dependencies resulting in a model that is low-dimensional, fast to train, and suitable for characterizing pedestrian behavior. 

Trajectory classification is a natural choice for quantifying actions in a space; however, it is difficult to determine the class labels a priori and annotation is impractical and time-consuming. Unsupervised and self-supervised methods for clustering have the advantage of requiring no labels. 
Recent work in unsupervised guidance for pre-training  builds convnets in a self-supervised manner, creating a network that emulates the output of the unsupervised tasks. For example, \cite{gidaris2020learning, gidaris2020online} learns a representation that matches visual bag-of-words output; \cite{xue2018deep} trains a network to match the 2D t-SNE output creating a latent space for texture recognition; \cite{caron2018deep, caron2020unsupervised} trains networks to generate clusters for unsupervised feature learning; \cite{tokmakov2020unsupervised} trains a 3D convnet to match the k-means clustering of an embedding space. 
% other papers ...
% Joint Unsupervised Learning of Deep Representations and Image Clusters
% Online Deep Clustering for Unsupervised Representation Learning
% clustering-based representation learning methods as a subset of unsupervised representaion learning
We follow this trend of using unsupervised methods to train clustering networks to build a t-SNE-imitator network for projecting pedestrian paths to a trajectory latent space. Moreover, we show that clusters in this latent space correspond to distinct and interpretable pedestrian behavior and therefore can be used to build a pedestrian behavior dictionary.

\section{Methods}
 
% \textbf{so the reason I put the trajectory math formulation first was bc that paragraph is a summary of all of our methods/applications that I put as a preamble to the actual section. I think launching right into talk of manipulating trajectories before you make sure the reader is on the same page about exactly what kinds of trajectories we're talking about is unwise? or even if we don't introduce it with math, maybe moving the first paragraph from the first section to this space will be sufficient to accomplish this goal?}

%\textbf{and in subsequent sections we describe three ways we use the dictionary; or make this the end of an intro paragraph to methods???} 
% multi-person trajectory into 2D latent space w/ like trajs together
% good utility bc can see how behavior clusters, make dictionary, 
% then apply to 3 applications: simple pred, space char., behavior char. (?) based on distances to find typical behaviors/anamalous behaviors 

% a semantically meaningful, social behavior cluster, $c^i$, that each pedestrian takes in that time frame. Additionally, we create pedestrian behavior dictionary from the collection of unique social behavior clusters found in the data. The trajectory predictions are conditioned on the social behavior clusters from the P-SAM to allow us to use much simpler networks for our predictions than SOTA methods. This mapping is also used to create a social heatmap of the scene and to find social waypoints, or points of social behavior inflection or change in the environment, in order to characterize and find patterns in how the space is utilized. 

\subsection{PT-net for a Trajectory Latent Space}
%\textbf{use t-SNE imitator network? instead of student teacher?}

%\textbf{work this in somewhere and add the seconds amount too!}
%Given raw trajectories $X^i_t = \{x_t^i, y_t^i\}$, for person $i$ \textbf{but this is for more than just 1 person. this is over all people? idk}in a scene over $t = \{1,2,...,T_{obs}\}$ time steps, we predict %both future trajectories, $ \hat{X}^i_t = \{\hat{x}^i_t, \hat{y}^i_t\}$, for times $t = \{T_{obs}+1, ..., T_{max}\}$ and 

%\textit{Need a lead in to discuss the overall steps. Something like: 
We devise PT-net to act as a t-SNE-imitator
%, without the instabilities of t-SNE,  
by using a student-teacher framework to obtain a stable %t-SNE  
embedding from multi-pedestrian trajectories, which are %. Trajectories are 
particularly well-suited for t-SNE-embedding because of their low dimensionality.
%Before computing the embedding, we process the raw trajectory data. 
Given trajectories $X^i_t = \{x_t^i, y_t^i\}$ for each person $i$ in a scene over $t$ time steps% = \{1,2,...,T_{obs}\}$ time steps, where $T_{obs}$ is the total number of observed time steps, 
, we split the trajectories into overlapping segments of length $T$ where each segment is offset by $\Delta T$, and each pedestrian is present for the duration of the time steps in each trajectory segment. 
This provides environment specific pedestrian behavior examples that are limited by features such as sidewalks, entryways, and roads. The trajectory segments are augmented by rotation ($\theta = 30, 45, 60 $ degrees) to insert synthetic variation into the %input trajectories 
data to make the learned latent space more generalizable. %less dependent on the observation region.
\begin{figure*}
    \centering
    \includegraphics[width=\textwidth]{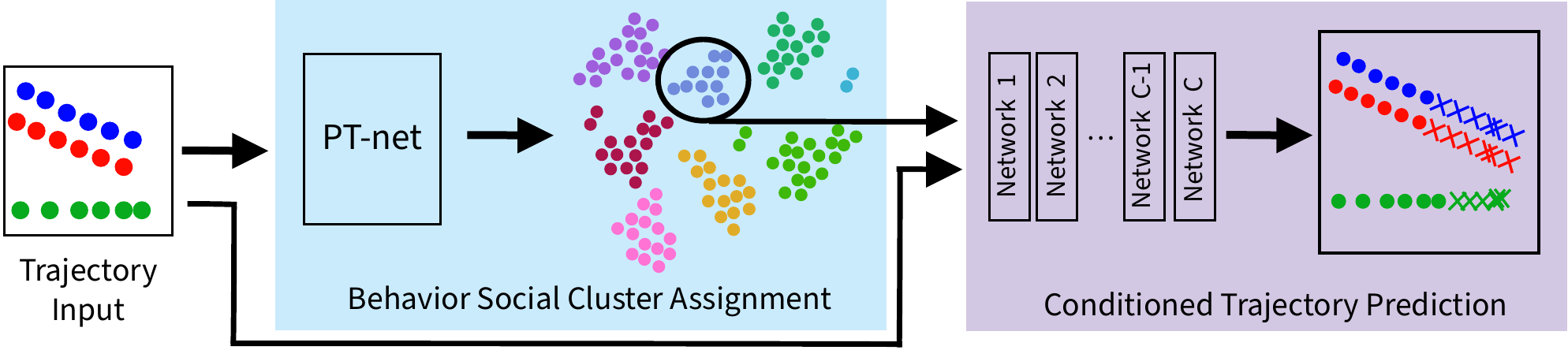}
    \caption{For pedestrian trajectory prediction, PT-net predicts the social behavior cluster assignment corresponding to the behavior of the pedestrians in the scene. This assignment dictates which of the MLPs will be used to predict the future trajectories of the pedestrians. We train one MLP per cluster in the pedestrian behavior dictionary and deterministically condition the prediction upon the social behaviors of the pedestrians in the scene. }
    \vspace{-10pt}
    \label{fig:supPredTask}
\end{figure*}

%
%To create the input features for the t-SNE algorithm
%
The relative velocity, $v^i$, is computed as follows:%, for each pedestrian in the trajectory segment, % from the trajectory segments in each group,
\vspace{-3pt}
\begin{equation}
    v^i= [x_{t}^i-x_{t-1}^i, y_{t}^i-y_{t-1}^i] %~~~ \forall ~t \in \{1,2,...,T_{obs}\}
    \vspace{-2pt}
\end{equation}
where $x^i_t$ and $y^i_t$ are the $(x,y)$ location of pedestrian $i$ at time $t \in T$.
Subsequently, $d^i$, the distance between person $i$ at time $t$ and the $N-1$ nearest pedestrians in the scene at time $t-1$, is computed for each time step in the segment, %for time steps $t \in [2, ...,T_{obs}]$, 
\vspace{-1pt}
\begin{equation}
    d^i = [x_t^i-x_{t-1}^j, y_t^i-y_{t-1}^j] ~~~ \forall ~j \in N \text{ where } j \neq i 
    \vspace{-1pt}
\end{equation}%~t \in \{2,...,T_{obs}\}, ~~
where $N$ is the total number of people in the trajectory segment. These two vectors, $v^i$ and $d^i$, are computed for each person in the trajectory and concatenated to form % before being used as the input to the t-SNE algorithm,
\vspace{-3pt}
\begin{equation}
    D = [\alpha v^i ~|~ d^i] ~~~ \forall ~ i \in N
    \vspace{-2pt}
\end{equation}
where $\alpha$ is a scaling factor included to combat the difference in scale between the relative velocities $v^i$ and the proximity-based features $d^i$ and $|$ denotes the concatenation operation. This process is repeated for each trajectory segment collected from the raw data. 

From there, we split the data into groups containing equal numbers of pedestrians  %based on the total number of pedestrians present in each scene 
and use t-SNE to create a trajectory embedding for each group. The t-SNE embedding output is clustered using k-means, where $k$ is chosen through visual inspection of the t-SNE embedding manifold. The cluster assignments of each point in t-SNE space are paired with their corresponding raw trajectories. Sampling small numbers of these points per cluster and comparing the associated raw trajectories reveals a taxonomy of semantically meaningful behaviors like leader-follower, walking in pairs, or standing around in small groups, as shown in Figure \ref{fig:t-SNEEmbedding}. %that are enumerated in Table \ref{tab:behaviorTaxonomy}. 
See Figure \ref{fig:process} for an overview of this process and Table \ref{tab:behaviorTaxonomy} for a textual description of the observed behaviors. %*** expound upon this figure reference?***

\begin{comment}
% I'm consolidating this and moving it to the intro
It may seem as if this discrete set of social behaviors could be defined manually, but this is not the case. For example, the leader-follower behavior appears simple at first glance. It comprises two pedestrians traveling approximately the same path, one in front of the other, separated by a distance threshold. However, this distance threshold varies between environments and pedestrian walking speed preferences. It is also difficult to define specific bounds to capture what ``approximately the same path" would look like because rigid thresholds would not allow for slight trajectory adjustments due to environmental changes, i.e.\ other pedestrians or vehicles in the scene. The benefit of our method is that these behaviors are clustered automatically without the explicit definition of these properties. Additionally, the dictionary allows for the discovery of these properties for a specific environment. 
\end{comment}

% \textbf{I can add the local meandering (I'll look for a better term for that) behavior example here too, but I think one might be enough?}

This trajectory embedding learned directly from the pedestrian velocities using t-SNE is useful, but t-SNE embeddings are irreproducible, which is %change each time they are computed, making them irreproducible. This is 
undesirable in the event that it is necessary to predict the social behavior clusters of previously unseen pedestrian trajectories or add to the number of social clusters by processing new data. To combat this, PT-net, a feed-forward MLP network, learns to mimic the resulting t-SNE embeddings. %for each $N$. 
% \textit{A modified and shortened version should go in the intro when we are defining terminology and stating what we are doing. I wouldn't use the term "regular behavior",  But we do need to define behavior. The clusters of multi-pedestrian trajectories correspond to human-interpretable {\it pedestrian behavior} in many cases and can be assigned semantic labels. For example,  as shown in Figure ??,  observable pedestian behaviors in the latent space clusters include (give 3 examples).  Note: Figure?? should show the latent space clusters and then an arrow to examples of multi-person trajectories and the figure should show the 3 examples named in this sentence.}
%PT-net learns the embedding of social behavior clusters, in an unsupervised, data-driven manner. 
Each cluster in the pedestrian behavior embedding corresponds to one unique social behavior. This effectively quantizes a continuum of behaviors to reduce the dimensionality of an infinite set of social behaviors, 
thus providing a tractable lexicon of social behaviors for high level scene reasoning.

\subsection{Utilizing the Pedestrian Behavior Dictionary}
\paragraph{\bf Trajectory Prediction with PT-Net}~

%Given raw trajectories $X^i_t = \{x_t^i, y_t^i\}$, for person $i$ \textbf{but this is for more than just 1 person. this is over all people? idk}in a scene over $t = \{1,2,...,T_{obs}\}$ time steps, 
\noindent Using a set of trajectories $X$ as input, we predict future trajectories $\hat{X}$ % = \{\hat{x}^i_t, \hat{y}^i_t\}$, for times $t = \{T_{obs}+1, ..., T_{max}\}$, %\textbf{say how long the horizons are and that it's following prior work and then cite the prior work as proof; change the lettering to whatever they're } where $T_{pred}$ is the prediction horizon.
deterministically conditioned on the predicted, semantically meaningful, social behavior cluster, $c^i$, from PT-net. Figure \ref{fig:supPredTask} illustrates this process. First,   the velocity and proximity input $D$ from Equation 3 is assembled for the current scene. %First, we make $D$ for the current scene from the raw trajectories. 
PT-net uses this to predict the social behavior cluster from the pedestrian behavior dictionary. Based on this cluster assignment, a specific feed-forward MLP network is chosen from an ensemble to make the pedestrian trajectory prediction. We train one MLP to predict future trajectories per cluster in the pedestrian behavior dictionary. This is possible because the dictionary effectively limits the possible state space of the trajectory prediction problem to a manageable range. 

%%% things to add to this section:
%   - this is a space specific methodology and we can make the simplification because we're modeling a specific space and train/test on same environment is a common thing done in cv/robotics/etc

\vspace{-10pt}
\paragraph{\bf Social Behavior and Environmental Characterization}~

%*** tighten this paragraph up! maybe make multiple ***
\noindent Once the pedestrian behavior dictionary is created, it can be used to answer questions about human behavior and environment utilization. A particular environment can be characterized by  enumerating the social behaviors that occur and computing the frequency of these behaviors. That is, pedestrian behavior histograms and pedestrian behavior maps allow  characterization of space usage and discovery of pedestrian behaviors (see Figures~\ref{fig:overlaps} and \ref{fig:histograms}). 
% While we cannot ensure that we've found all the distinct social behaviors, this work can still facilitate research in the fields on psychology, cognitive science, public policy, and beyond. 
Analysis of space in this manner is directly applicable to social science issues such as  public space assessment \cite{whyte1980social, whyte2015design, small2019role, honey2021impact} and pedestrian behavior analysis \cite{kothari2021interpretable,antonini2006behavioral,robin2009specification}. Key questions in pedestrian behavior can be answered such as:  what is the average distance in leader-follower relationships? Is this distance culturally dependent? What is the radius of movement when people are congregating during conversation or waiting?

\section{Results}

\subsection{Datasets}

We test our networks against SOTA methods using the ETH \cite{ETH} and UCY \cite{UCY} datasets on the task of pedestrian trajectory prediction. After creating the pedestrian behavior dictionary from the combination of trajectories from both datasets, PT-net is used to condition the trajectory prediction. For this task, the trajectories are normalized based on the scene sizes and centered on the origin so that the same networks can be used over multiple datasets. Note that each of the datasets occur on sidewalked areas. This limits the types of  behaviors that pedestrians can exhibit. For example, it is highly unlikely that a pedestrian will walk in circles or meander directionlessly because the social convention is to walk parallel to the buildings. Subsequently, the trajectories are augmented with several rotations %before creating the pedestrian behavior dictionary 
to introduce more behavior variety. 

\subsection{Training}

\paragraph{\textbf{Behavior Dictionary Cluster Prediction}}~

\noindent To create the input data for the t-SNE, we choose sliding windows of size $t=8$ time steps and $\Delta t=1$ for ease of comparison with the experiments in SOTA and to learn behaviors that would be relevant to existing work. The sliding windows of trajectories allow for more granular detection of social behavior changes. We choose $\alpha$ to be 15, and the 

\begin{figure}[H]
    \centering
    \includegraphics[width=.48\textwidth]{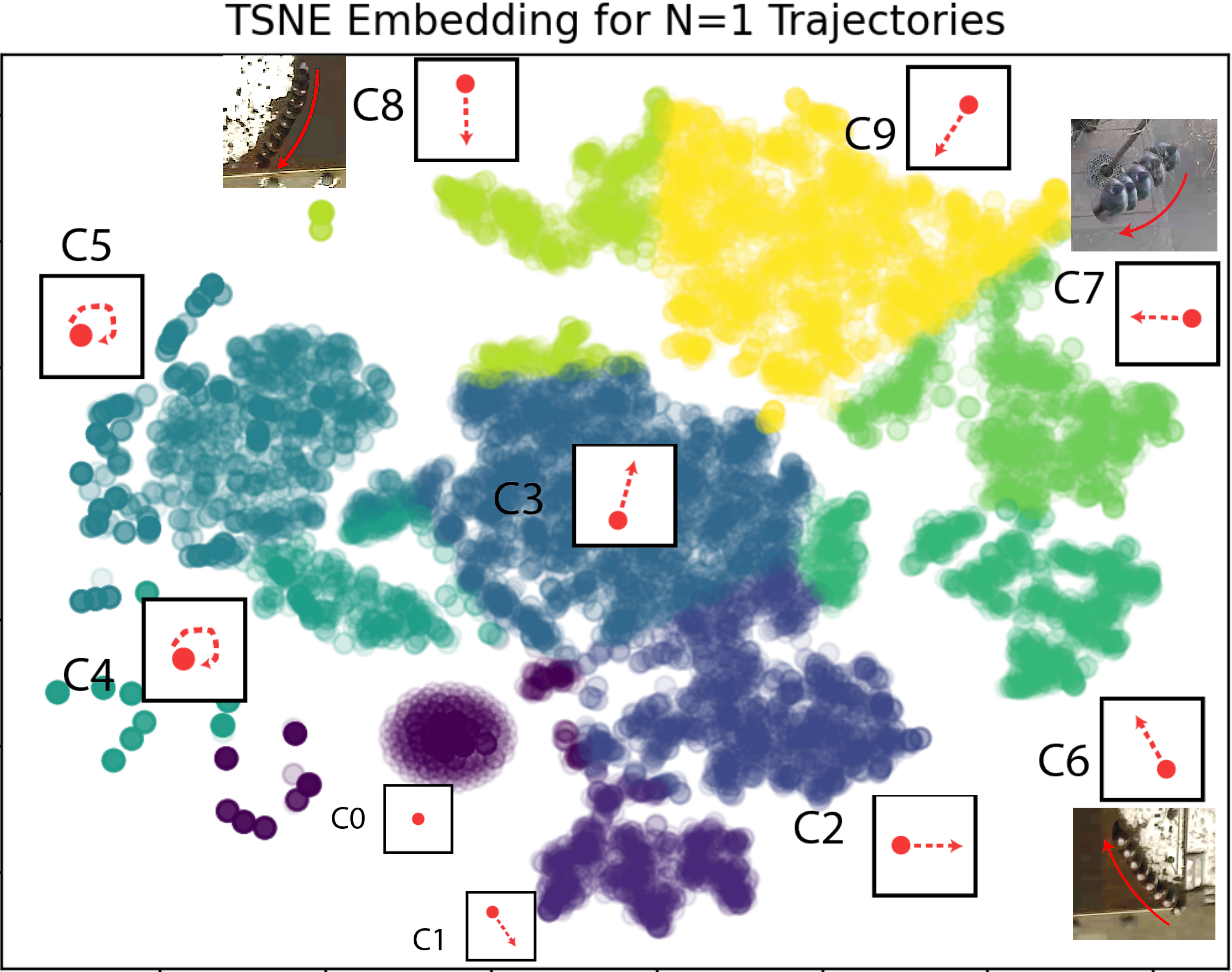}
    \includegraphics[width=.48\textwidth]{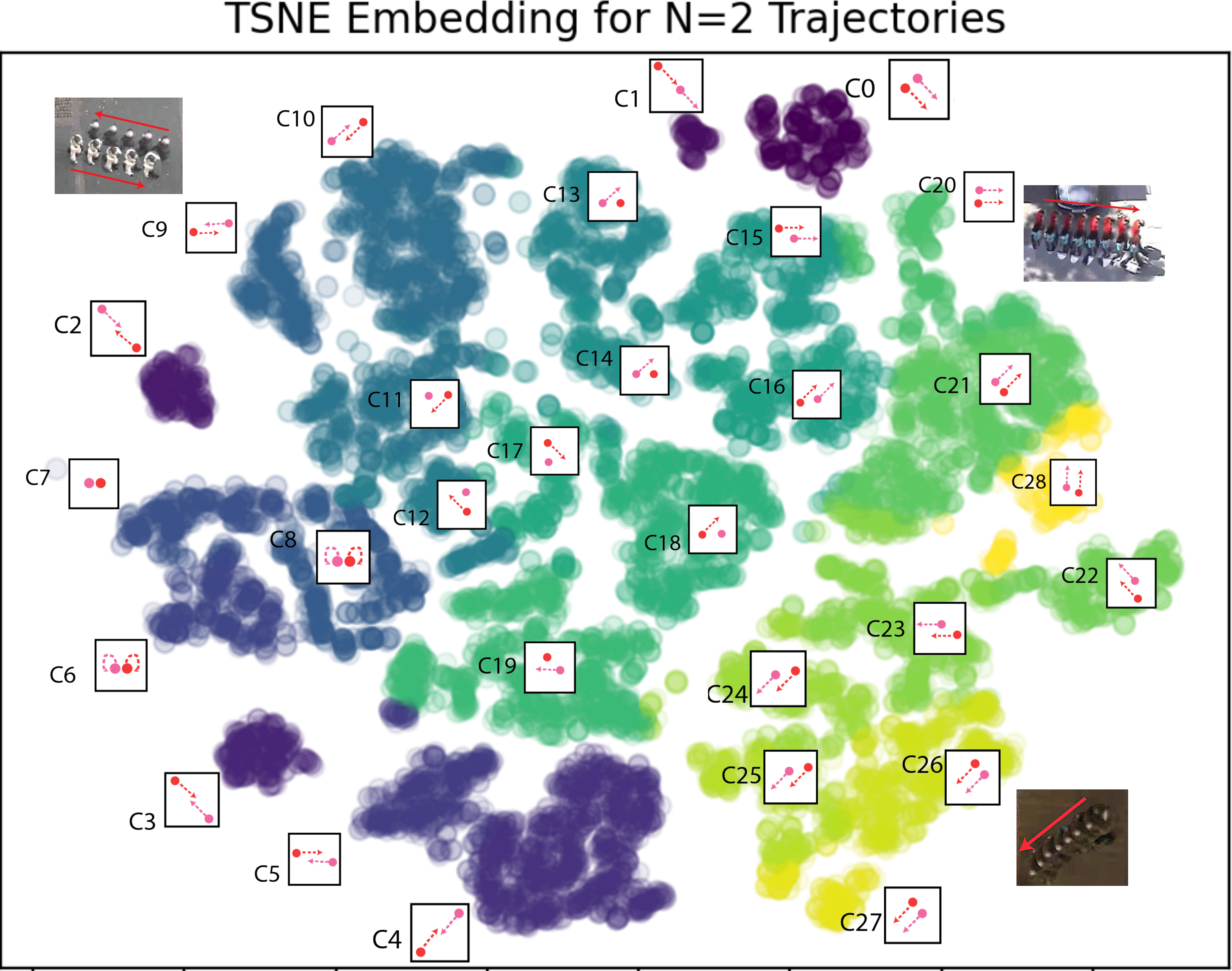}
    \includegraphics[width=.48\textwidth]{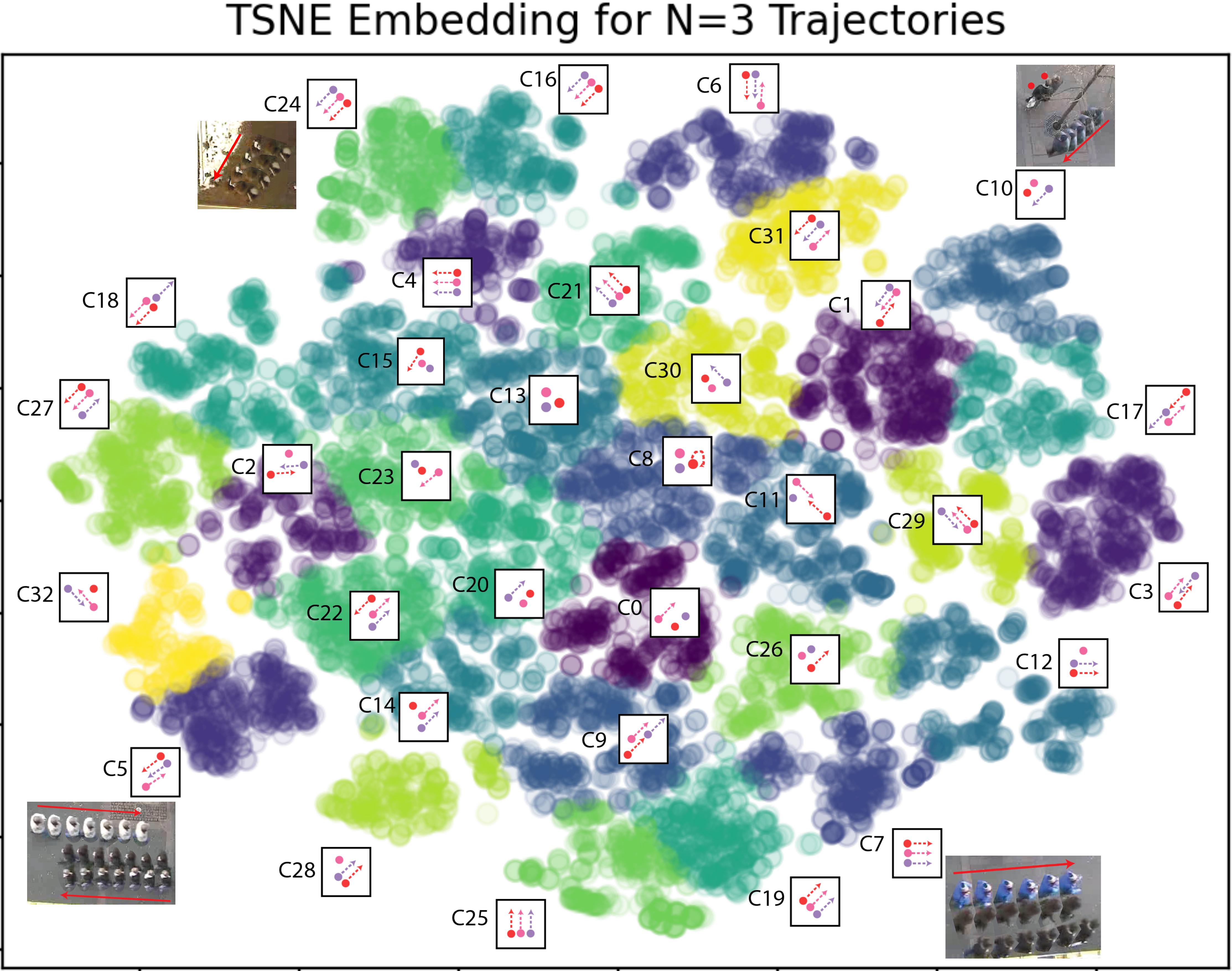}
    \caption{Each dot represents the t-SNE embedding of a 3.2 second trajectory. The colored clusters denote distinct social behaviors for N=1,2,3 people (corresponding colors across the three graphs do \textbf{not} denote related behaviors) along with trajectory diagrams, showing simplified sketches of the behaviors from Table \ref{tab:behaviorTaxonomy}. }
    \label{fig:t-SNEEmbedding}
\end{figure}

\begin{figure*}[h]
    \centering\captionsetup[subfloat]{labelfont=bf}
    \begin{subfigure}[b]{0.33\textwidth}
        \includegraphics[width=\textwidth]{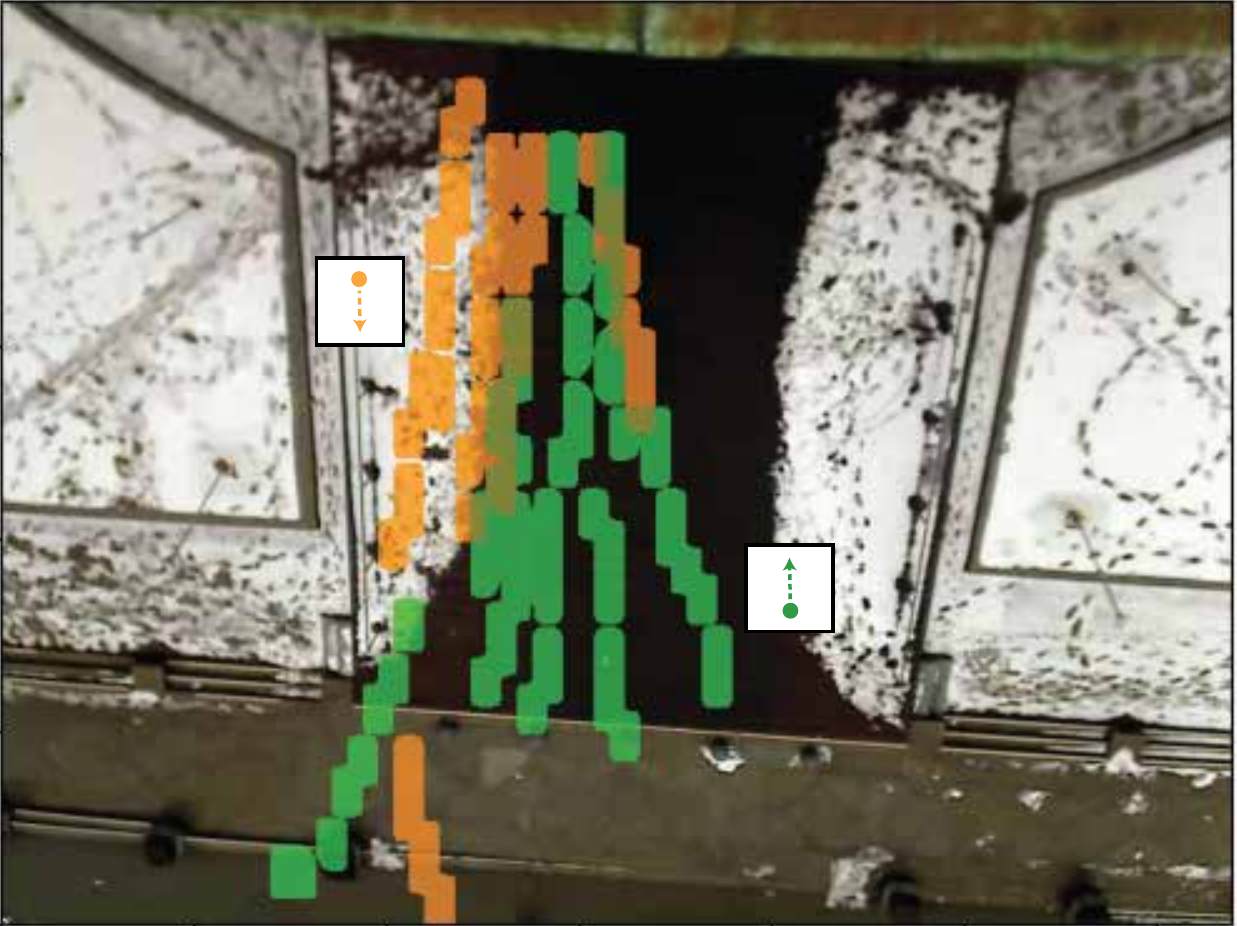}
        \caption{N=1, Clusters 1 and 9}
        \label{fig:N1C1/9Heatmap}
    \end{subfigure}
    \begin{subfigure}[b]{0.33\textwidth}
        \includegraphics[width=\textwidth]{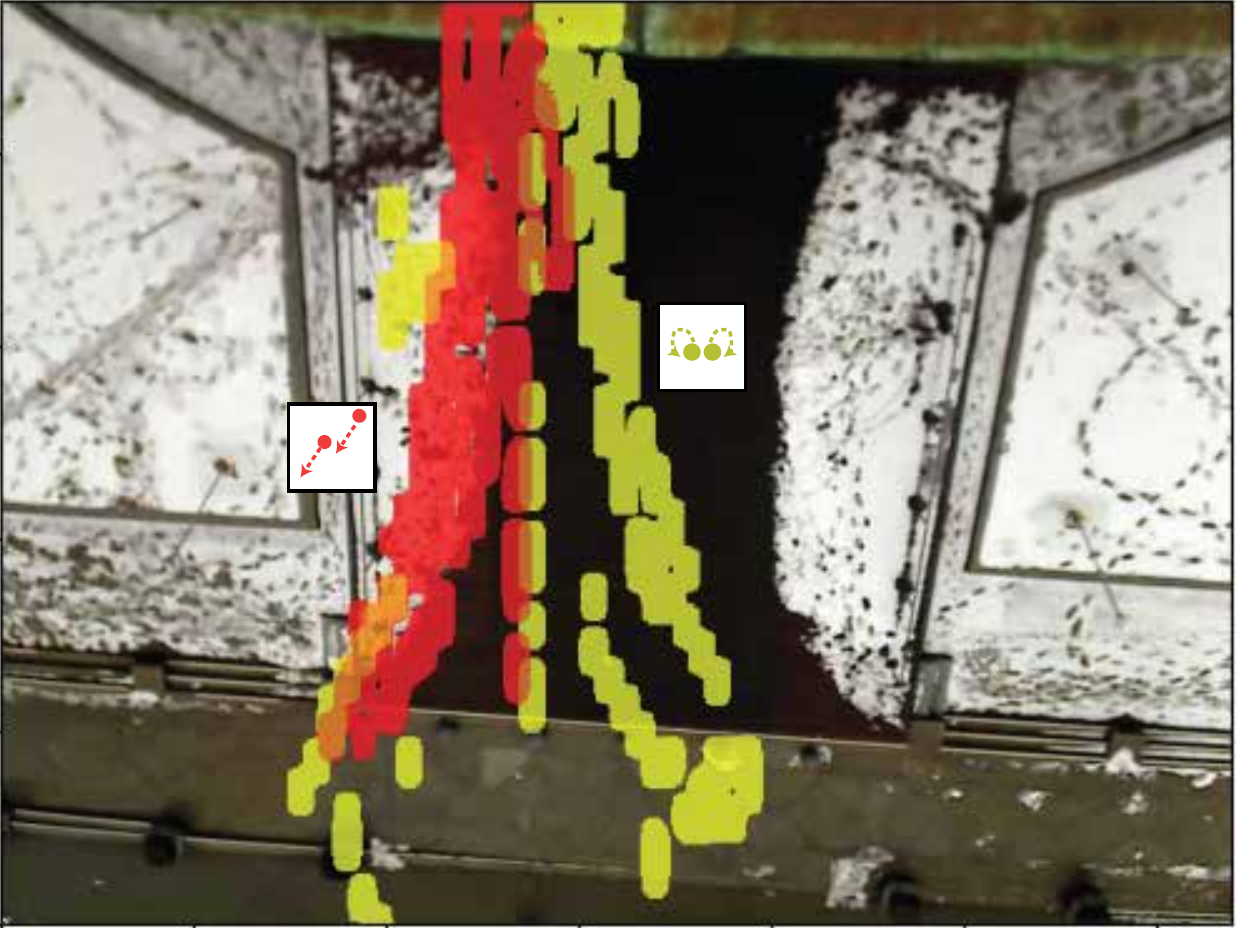}
        \caption{N=2, Clusters 6 and 22}
        \label{fig:N2C6/22Heatmap}
    \end{subfigure}
    \begin{subfigure}[b]{0.33\textwidth}
        \includegraphics[width=\textwidth]{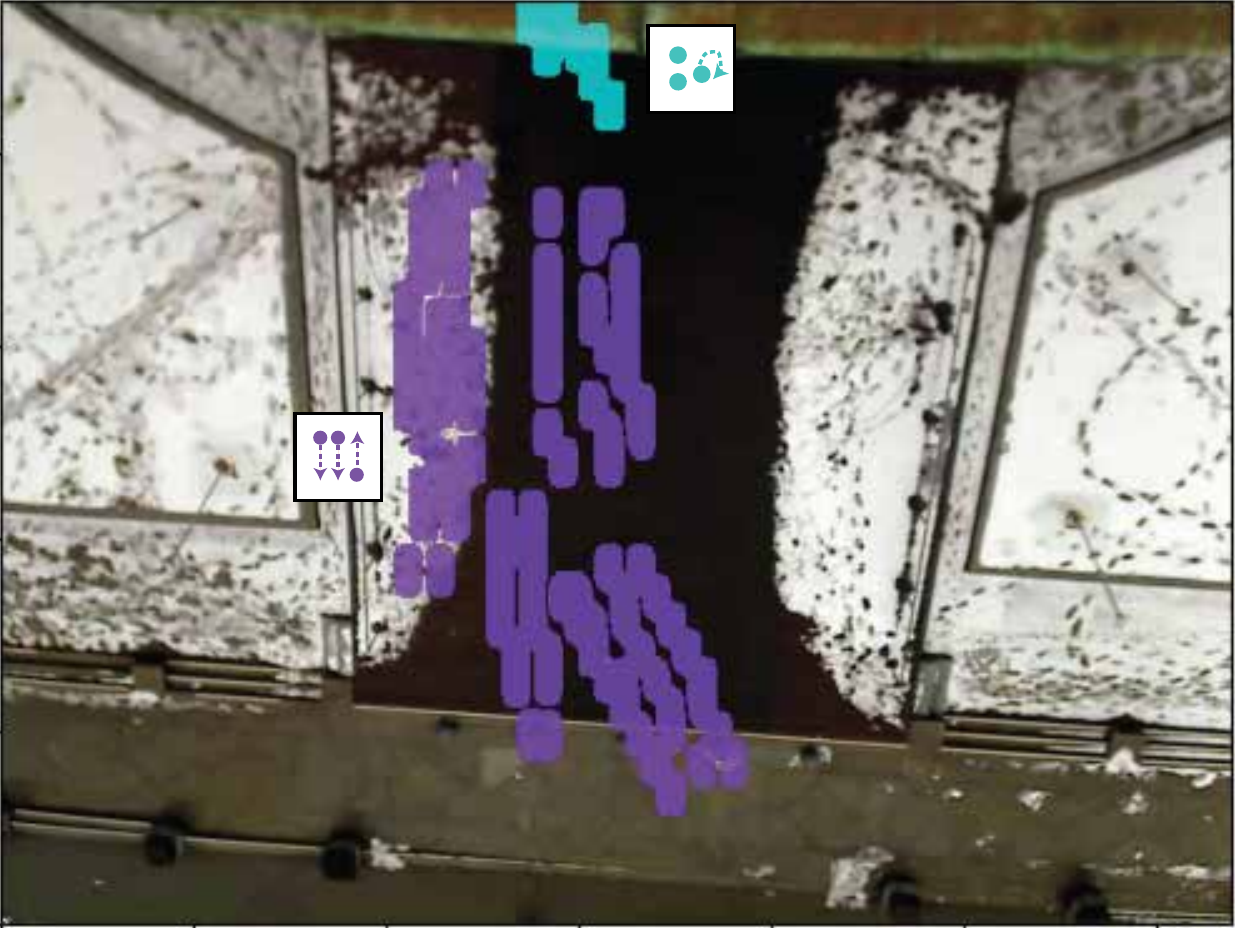}
        \caption{N=3, Clusters 5 and 8}
        \label{fig:N3C5/8Heatmap}
    \end{subfigure}
    
    \caption{
    {\bf Pedestrian Behavior maps}.  Colored boxes indicate select pedestrian behaviors occurring over the entirety of the data collection period for ETH. 
    We can infer a rich story about environment usage for varying  numbers of pedestrians (N) and social behavior clusters (C). \textbf{(\ref{fig:N1C1/9Heatmap})} (Green) Person entering the building; (Orange) Person leaving the building. Notice that people leave the building in a more narrowed and constrained path, indicating they are giving right-of-way to those entering.
    \textbf{(\ref{fig:N2C6/22Heatmap})} (Yellow) Two people standing still together; (Red) Two people leaving the building to the left in a leader-follower formation. Notice that the people standing still tend to congregate off to the sides, or at an island in the middle of the walkway, and the people exiting fit into the gaps left behind. 
    \textbf{(\ref{fig:N3C5/8Heatmap})} (Purple) Two people walking side by side to exit the building passing one person entering the building; (Blue) Three people standing still. Notice there is a bottleneck around the door that prevents pedestrians from moving, but it becomes easier to move freely further from the door.}
    \vspace{-5pt}
    \label{fig:overlaps}
\end{figure*}

\begin{table}[]
    \footnotesize
    \centering
    \begin{tabular}{| p{5.5cm} | p{2cm} |}
        \hline
         \multirow{2}{5.5cm}{Semantic Behavior} & \# People (N),\\
         &  Cluster \# (C) \\
         \hline
         \hline
         \multirow{2}{5.5cm}{Standing Still} & N1:C0, N2:C7\\
         & N3:C13\\
         \hline
         \multirow{1}{5.5cm}{Walking Straight}  & N1:C1-3, 6-9  \\
         \hline
         \multirow{2}{5.5cm}{Congregating} & N1: C4-5\\
         & N2:C6, 8\\
         \hline
         \multirow{3}{5.5cm}{Walking Side by Side} & N2:C0,20,21,26,27 \\
         & N3:C4,7, 16, 19\\
         & N3:C21, 24-25 \\
         \hline
         \multirow{2}{5.5cm}{Leader-Follower} & N2:C1, 15-16,\\
         & 22-25\\
         \hline
         \multirow{1}{5.5cm}{Two Passing in Opposite Directions} & N2:C2-5, 9-10 \\
         \hline
         \multirow{1}{5.5cm}{One Passing, One Standing} & N2:C11-14,17-19 \\
         \hline
         \multirow{2}{5.5cm}{One Passing a Pair Standing} & N3:C0, 10, 15,\\
         & 20, 23, 26, 30 \\
         \hline
         \multirow{2}{5.5cm}{Pair Passing One Walking in Opposite Direction} & N3:C1, 5-6, 22,\\
         & 27, 29, 31\\
         \hline
         \multirow{2}{5.5cm}{Two Walking in Opposite Directions Passing One Standing} & N3:C2, 32\\
         & \\
         \hline
         \multirow{2}{5.5cm}{One Walking Between Pair in Opposite Direction} & N3:C3\\
         & \\
         \hline
         \multirow{1}{5.5cm}{Two Still, One Fidgeting} & N3:C8 \\
         \hline
         \multirow{1}{5.5cm}{Two Leader-Follower, One Parallel} & N3:C9, 17 \\
         \hline
         \multirow{2}{5.5cm}{Two Passing Opposite Directions, One Standing} & N3:C11 \\
         & \\
         \hline
         \multirow{1}{5.5cm}{Pair Walking Past One Standing} & N3:C12, 14, 28 \\
         \hline
         \multirow{1}{5.5cm}{Pair Walking Away From One Walking} & N3:C18 \\
         \hline
    \end{tabular}
    \caption{A summary of the observed semantic behaviors for N=1,2,3 people with the corresponding cluster indices (C). Behaviors with different N or executed in different directions with the same semantic description are grouped together. Refer to Figure \ref{fig:t-SNEEmbedding} for a visual characterization of each behavior.}
    \label{tab:behaviorTaxonomy}
    \vspace{-8pt}
\end{table}

\noindent maximum number of people in each trajectory used for our experiments is $N=3$. In practice, there are much more than three pedestrians in the scene at a time. To deal with this, the larger groups are broken down into subgroups so that PT-net only predicts on the the closest $N$ pedestrians at a time. This process is repeated until each person is included in at least one group. PT-net is a collection of 3 MLPs, one for each N, trained for 300 epochs. Each MLP is comprised of four linear layers with ReLU activation functions and residual connections.

% \subsection{Social Behavior Heat Maps}
% The pedestrian behavior dictionary allows us to generate heat maps of the locations in the environment where pedestrians exhibit each social behavior. This gives further insight into how pedestrians use a space. 

% \subsection{Social Video Summary}
% Pairing the P-SAM with human feedback also allows us to automatically create video summaries of interesting events that take place in a certain area over a given period of time. 

\vspace{-10pt}
\paragraph{\textbf{Pedestrian Trajectory Prediction}}
For the trajectory prediction task, positions $X^i_t = \{x_t^i, y_t^i\}$ for each person $i$ in a scene over $t=8$ timesteps are input to the ensemble of MLPs which predict $ \hat{X}^i_t = \{\hat{x}^i_t, \hat{y}^i_t\}$ positions over $t=12$ timesteps into the future for each person. This choice of input and prediction horizons is standard in multiple SOTA methods\cite{gupta2018social,yuan2021agentformer,alahi2016social}. The ensemble networks are made up of four linear layers with the ReLU activation function and residual connections and are each trained for 1000 epochs on only the input data corresponding to their respective cluster assignments. %We compare our method to Social-LSTM \cite{alahi2016social} and D-LSTM \cite{kothari2021human} in Table \ref{tab:SOTAcompare}. 
Because PT-net learns space-specific features, we train one set of MLPs on 80\% (2880 trajectories) of the data from all scenes and test on the remaining 20\% (720 trajectories) for each environment. Training a scene-specific model inherently accounts for scene context. While some applications require no pre-training on the scene, the observation of a scene before algorithm deployment is quite reasonable in numerous applications, such as IOT, smart buildings, and traffic monitoring, due to a ubiquitous fixed camera.

\begin{figure*}
    \centering
    \includegraphics[width=\textwidth]{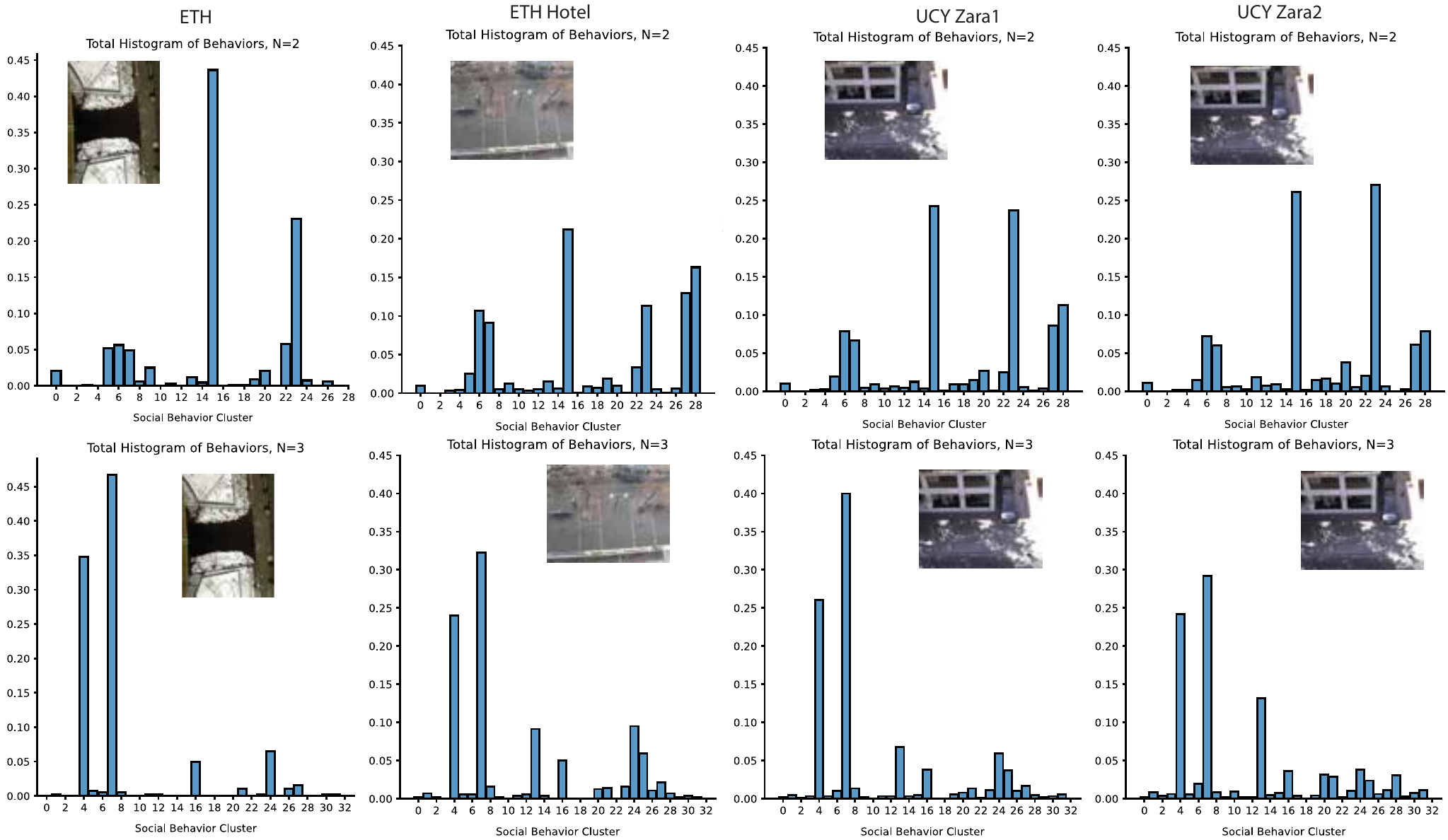}
    
    \caption{The histogram of behaviors in each environment (left to right: ETH, ETH Hotel, UCY Zara1, UCY Zara2) for $N=2,3$ people (top to bottom:N=2, N=3). From the histograms, it is evident that pedestrians utilize a different distribution of behaviors in ETH as opposed to all other environments. Because ETH depicts people walking in and out of a building, it is a much more constrained space than the open sidewalks in the other environments.  
    Even between UCY Zara1 and UCY Zara 2, which take place in the same environment at different times, there is still variation due to differing numbers of pedestrians and different pedestrian behavior patterns as the day progresses.}
    \label{fig:histograms}
    % more people walk updown in ETH (which looks like R/L bc of turn) bc sidewalks but in Hotel they have a large up/down (that's actually up/down) bc sometimes a train comes 
    \vspace{-5pt}
\end{figure*}

% \vspace{-15pt}
\subsection{Interpretable Pedestrian Behavior Dictionary}

The learned latent space for $N=1,2,3$ pedestrians is shown in Figure \ref{fig:t-SNEEmbedding}. Drawings with representative example trajectories are superimposed on each graph next to their corresponding clusters. Distinct clusters appear for each $N$ which indicates that our velocity and proximity-based trajectory processing is sufficient for learning distinct behaviors from the trajectories. Because we use velocity-based features, similar behaviors that are executed in different directions (ie. left-right vs top-bottom) often form different clusters. A description of the semantic behaviors associated with each cluster can be found in Table \ref{tab:behaviorTaxonomy}. The descriptions are condensed to only include the semantically distinct behaviors for each $N$. 

% table of cluster index - "manual cluster annotation" - semantic meaning - heatmap color (and combine similar semantic meanings ignore direction like leader follower - u/d/r/l) in a figure???? then list the full list of behaviors in the caption; for n=3 the behaviors require a longer text description (see supplemental material), some examples include... --> quantify manual labeling required

For $N$ larger than 3, it is still possible to find social behavior clusters, but their interpretability becomes more challenging as their number increases significantly with each subsequent increase of $N$ and behaviors become more complex. Behavior separation is smaller for high dimensions ($N=3$ in Figure \ref{fig:t-SNEEmbedding}); however, the embedding space still groups similar pedestrian behaviors in a sufficient manner for the downstream tasks of  predicting future pedestrian trajectories and interpreting behavior patterns in a space.

\subsection{Pedestrian Behavior Maps}

We use the pedestrian behavior dictionary to characterize space usage and social behavior patterns by creating behavior maps for each social behavior cluster in an environment. Clusters exhibiting the same behaviors with different pedestrian permutations are combined into the same behavior maps. For example in $N=2$, clusters $C=2$ and $C=3$ show the same semantic and directional behavior, but with the pedestrian order switched. The behavior maps of different clusters are superimposed to analyze the inter-pedestrian and inter-social group interactions that take place.
%It is important to note that it is impossible to infer causality of the behaviors from the behavior maps without estimating the pedestrian's theory of mind, but we can analyze their behavior to find interesting patterns in the space. 

Figure \ref{fig:overlaps} shows a selection of behavior maps in the ETH environment depicting the location that a particular behavior was exhibited over the duration of the entire dataset. Figure \ref{fig:N1C1/9Heatmap} shows the behavior map for one pedestrian entering the building (green) juxtaposed with one pedestrian leaving the building (orange). There are more people entering the building from the right, thus forcing the people exiting to stay to the left and give right of way to those entering. This could imply that there is a significant point of interest to the right of the building. 

Figure \ref{fig:N2C6/22Heatmap} overlays the behavior map of two people standing still together (yellow) and two people leaving the building leader-follower (red). People tend to stand still off to the sides, or at an island in the middle of the walkway, where they believe they will be most out of the way. The people leaving the building are forced to travel in the gaps between these congregators. Figure \ref{fig:N3C5/8Heatmap} combines the behavior maps of two people walking side by side to exit the building passing one person entering the building (purple) and that of three people standing still together (blue). The behavior map for three people standing still is localized around the entrance to the building, implying there is a bottleneck around the door   preventing people from moving freely. This bottleneck dissipates further away from the door where there is more free space. 

\begin{figure*}
    \centering
    \includegraphics[width=0.24\textwidth]{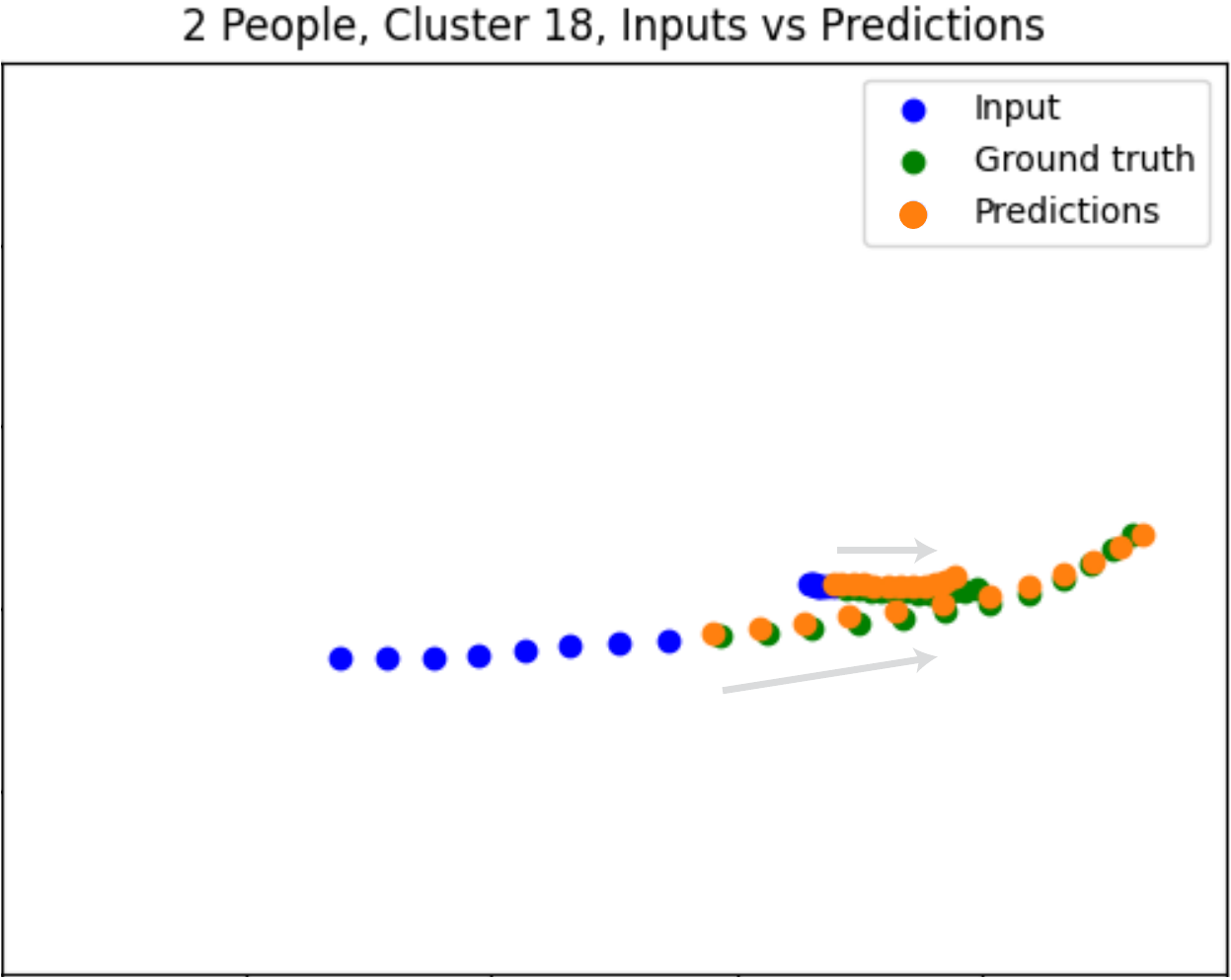}
    \includegraphics[width=0.24\textwidth]{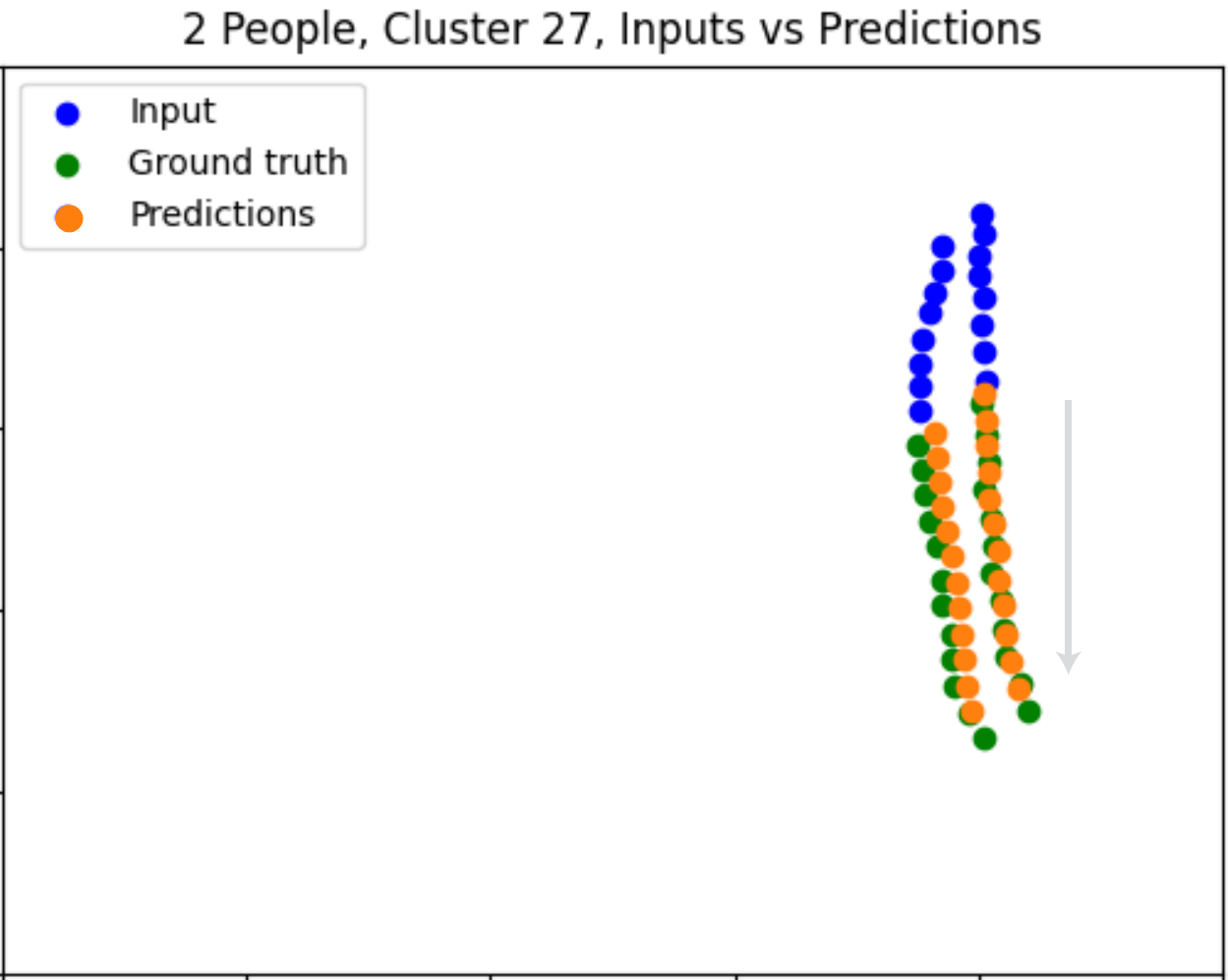}
    \includegraphics[width=0.24\textwidth]{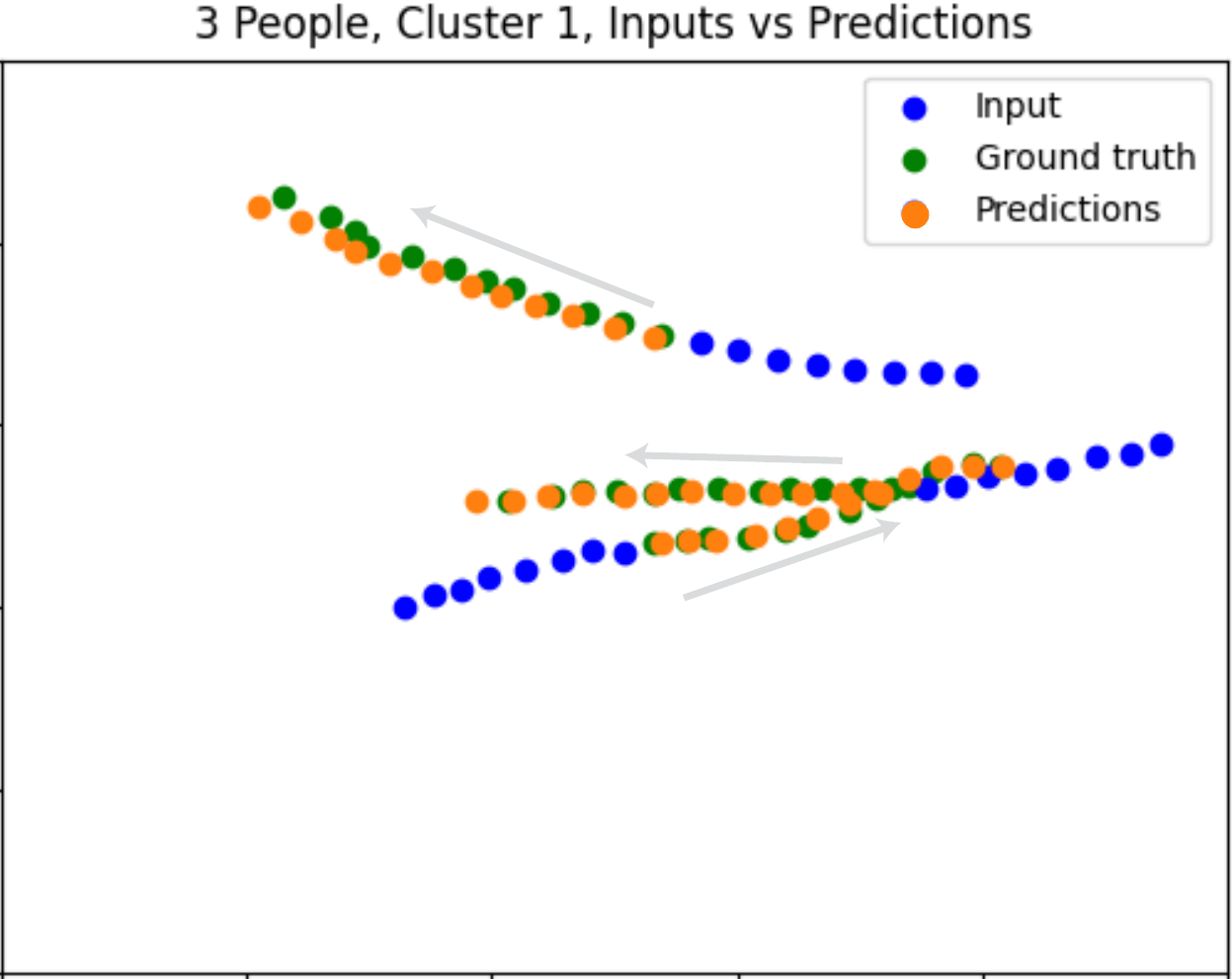}
    \includegraphics[width=0.24\textwidth]{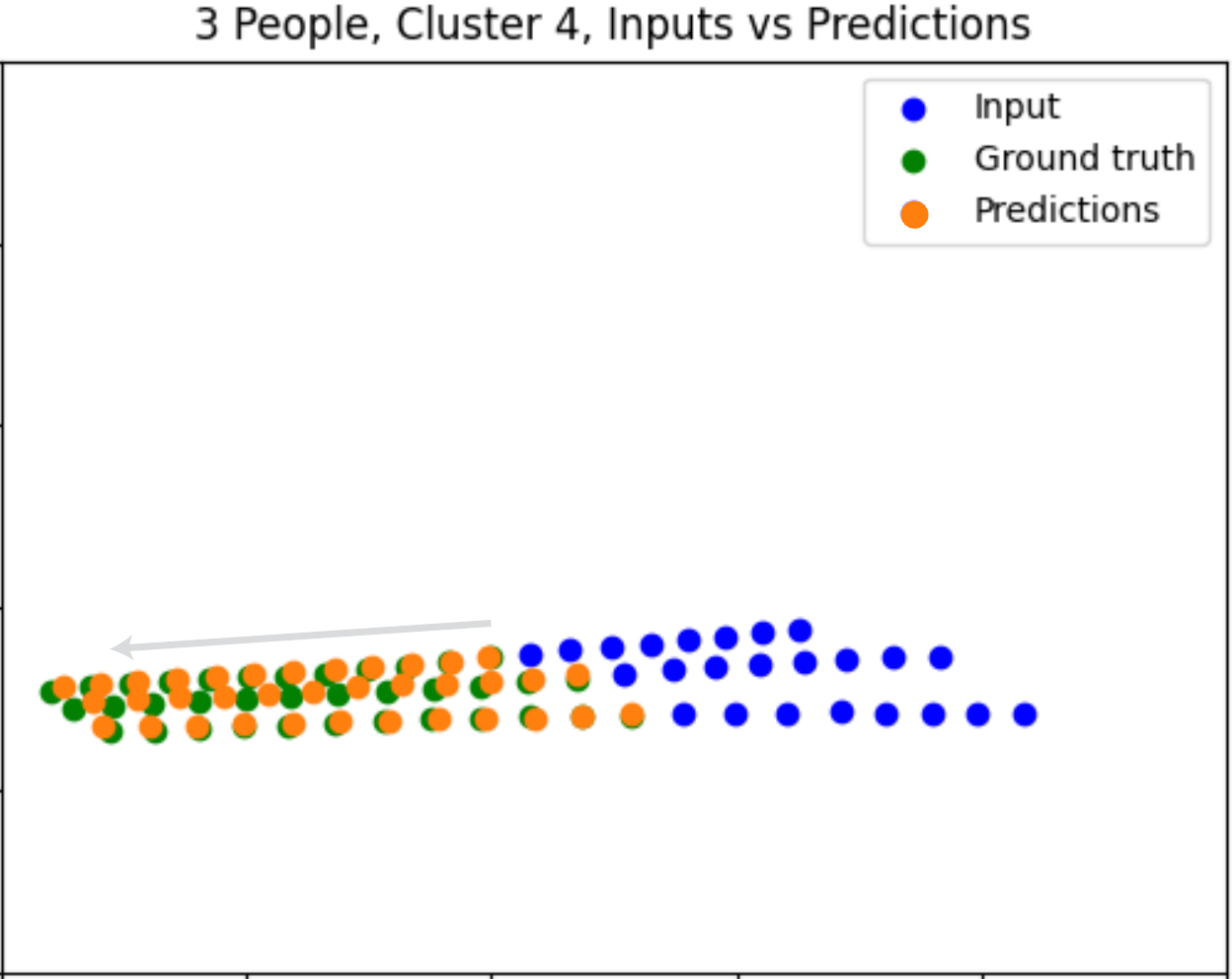}

    \includegraphics[width=0.24\textwidth]{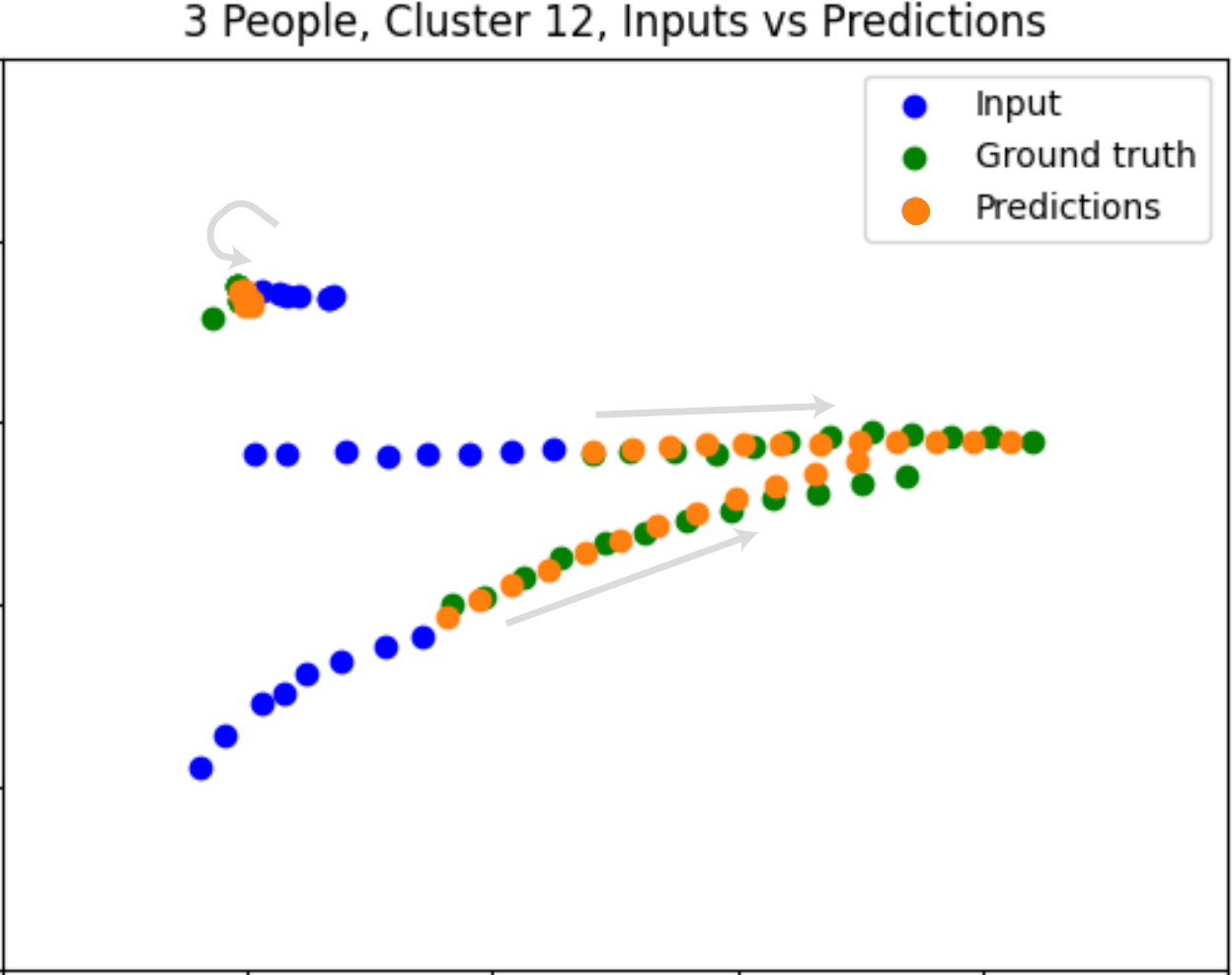}
    \includegraphics[width=0.24\textwidth]{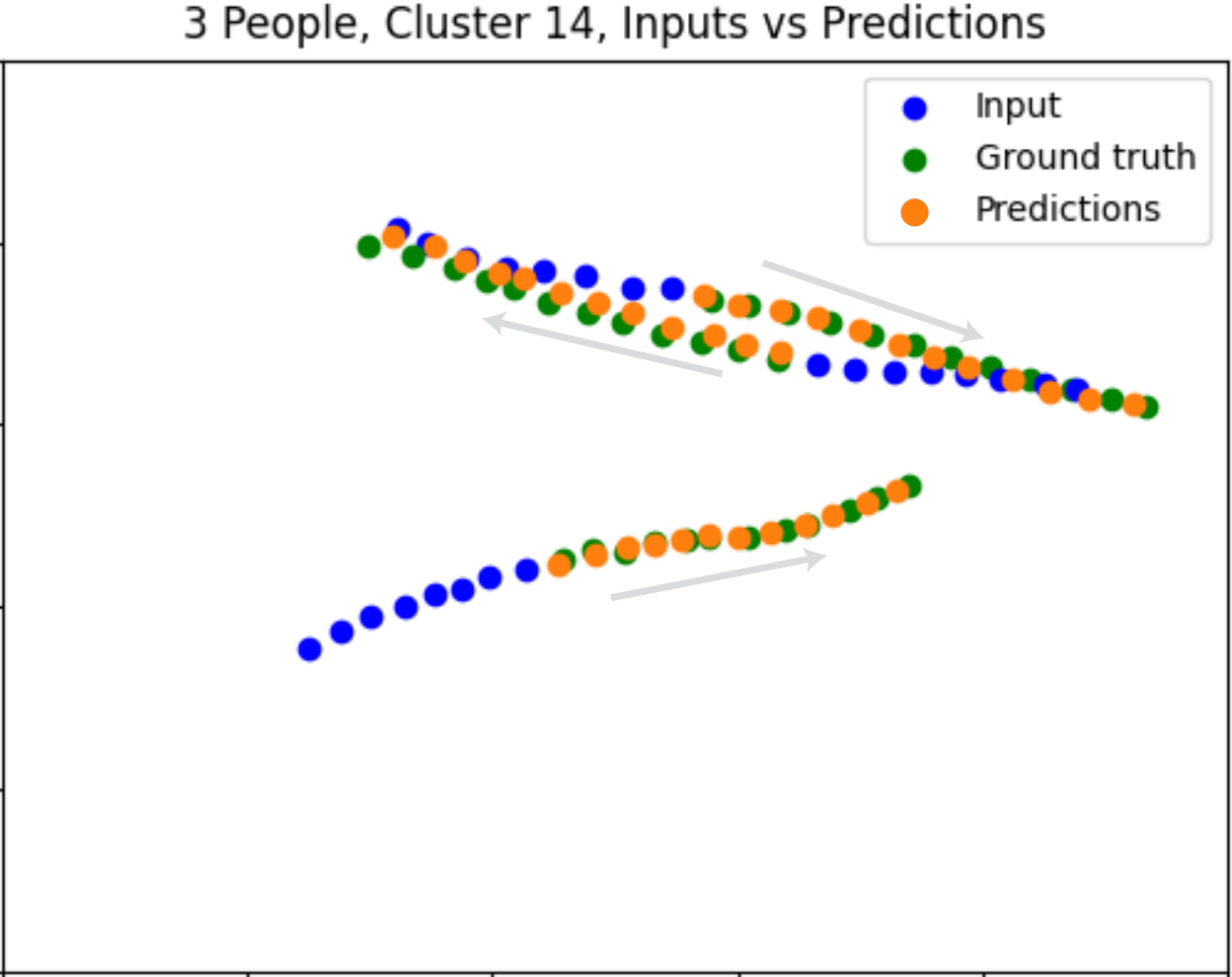}
    \includegraphics[width=0.24\textwidth]{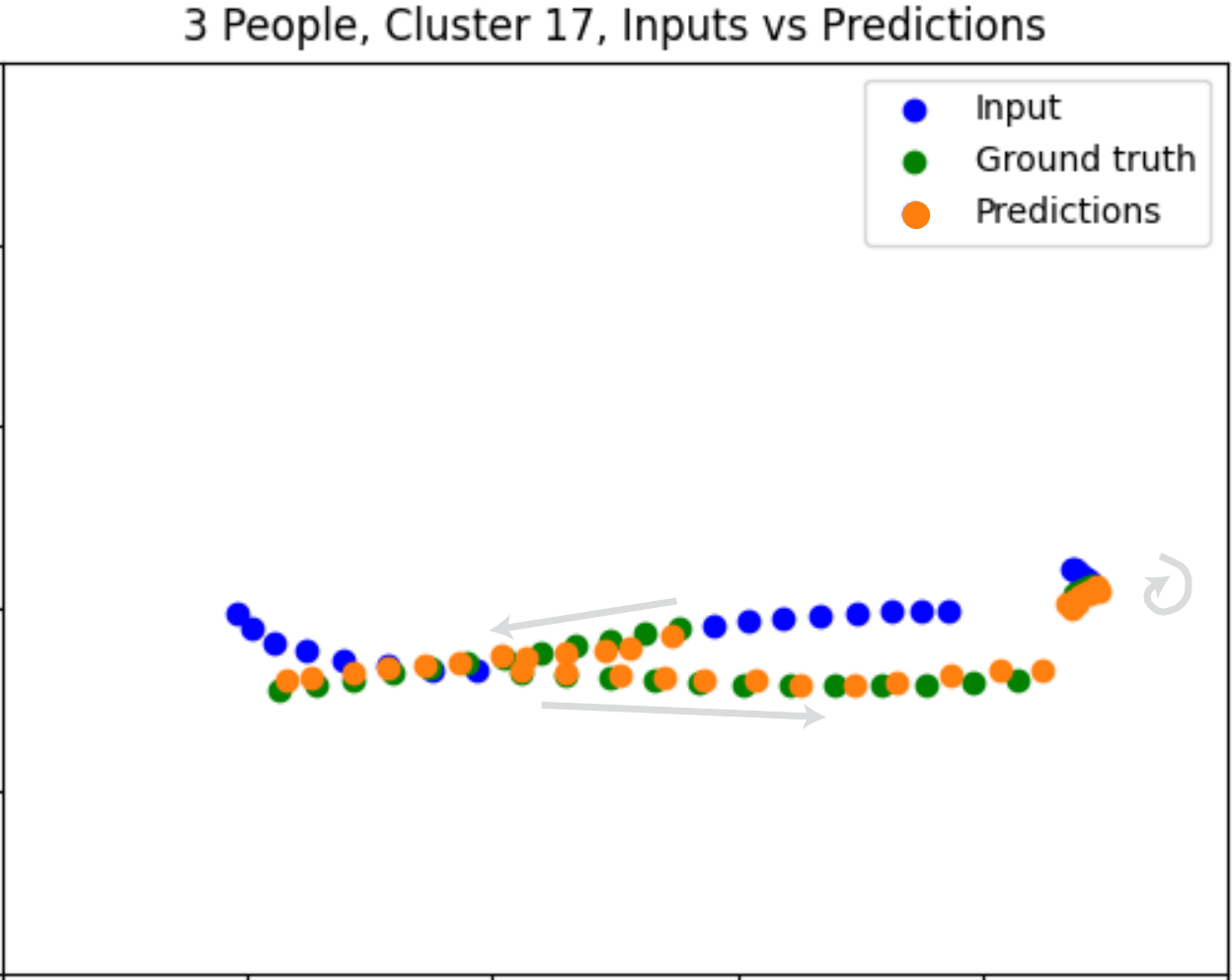}
    \includegraphics[width=0.24\textwidth]{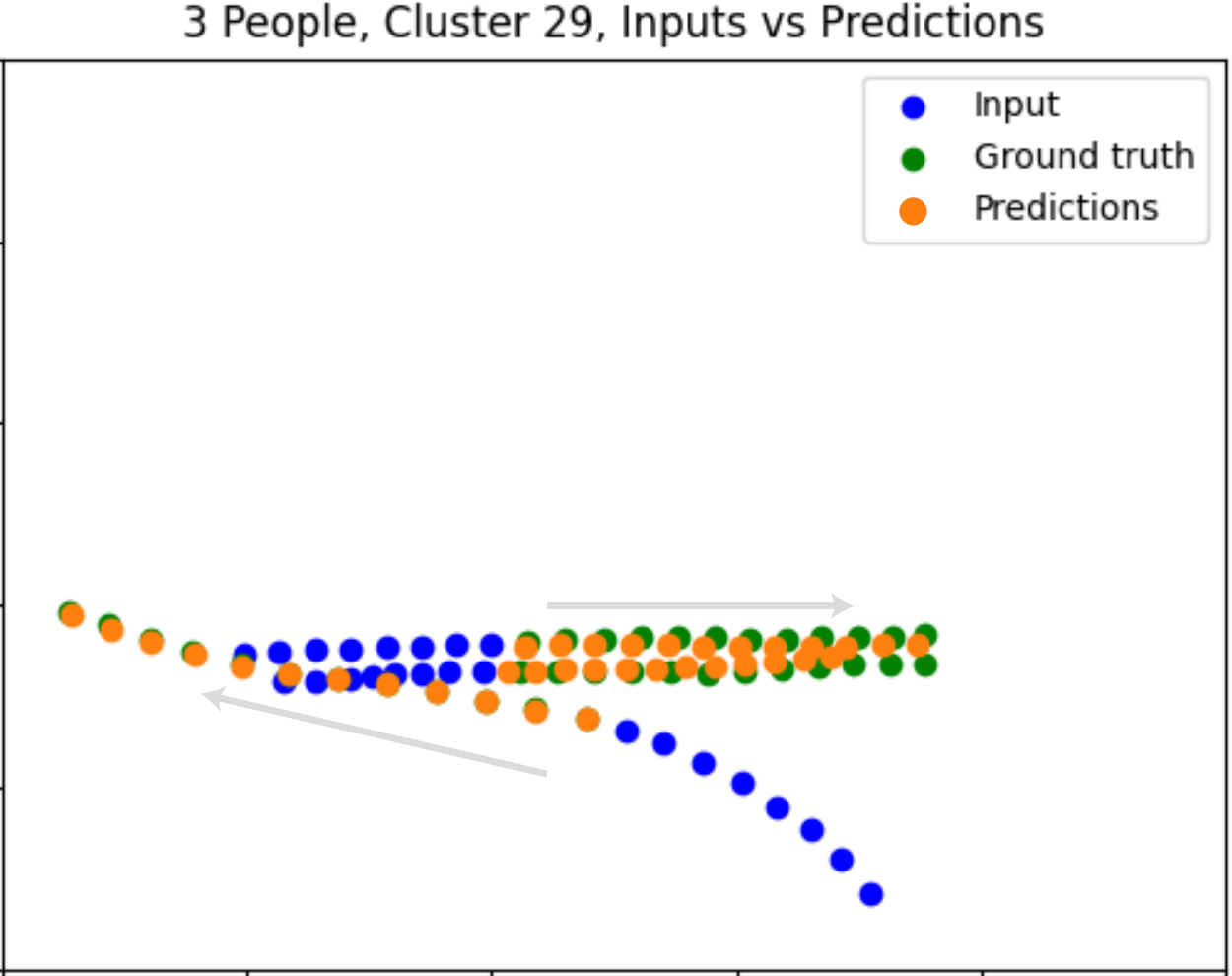}
    
    \caption{Example trajectory predictions from a selection of the predicted pedestrian behavior dictionary clusters. The input trajectories are in blue, the ground truth future trajectories are in green, and the predictions are in orange. The arrows in each diagram show the relative movement directions of the pedestrians. Our framework provides accurate trajectory prediction by conditioning on learned behaviors.}
    \label{fig:trajPerCluster}
    % \vspace{-5pt}
\end{figure*}

\begin{table*}[]
    % \hspace{-10pt}
    % \scriptsize
    \small
    \centering
    \begin{tabular}{|c|c|c|c|c|c|}
        \hline
        \multirow{2}{0.85cm}{Method} &  \multicolumn{5}{|c|}{ADE$_{20}$ / FDE$_{20}$ }\\
        \cline{2-6}
        & ETH & Hotel & Zara1 & Zara2 & Avg \\
        \hline 
        \hline
         SGAN\cite{gupta2018social} & 0.81/1.52 & 0.72/1.61 & 0.34/0.69 & 0.42/0.84 & .57/1.17\\
         Sophie\cite{sadeghian2019sophie} & 0.70/1.43 & 0.76/1.67 & 0.30/0.63 & 0.38/0.78 & 0.54/1.13 \\
         Transformer-TF\cite{giuliari2021transformer} & 0.61/1.12 & 0.18/0.30 & 0.22/0.38 & 0.17/0.32 & 0.30/0.53 \\
         STAR\cite{yu2020spatio} & 0.36/0.65 & 0.17/0.36 & 0.26/0.55 & 0.22/0.46 & 0.25/0.51 \\
         PECNet\cite{mangalam2020not} & 0.54/0.87 & 0.18/0.24 &  0.22/0.39 & 0.17/0.30 & 0.28/0.45\\
         Trajectron++\cite{salzmann2020trajectron++} & 0.39/0.83 & 0.12/0.21 & 0.15/0.33 & 0.11/0.25 & 0.19/0.41\\
         AgentFormer\cite{yuan2021agentformer} & 0.26/0.39 & 0.11/0.14 & 0.15/0.23 & 0.14/0.24 & 0.17/0.25\\
         \hline
         \hline 
         & \multicolumn{5}{c|}{ADE / FDE }\\
         \hline
         PT-Net (Ours) & 0.41/1.07  & 0.63/1.33 & 0.15/0.38 & 0.16/0.42 & 0.33/0.80\\
         \hline
    \end{tabular}
    \vspace{5pt}
    \caption{Comparison of the ADE/FDE of PT-net and SOTA for the ETH and UCY datasets. For SOTA methods, each network is trained with the leave one out strategy and tested on the remaining dataset with $k=20$ samples. PT-net learns scene-specific features, so it is trained on 80\% of the total trajectory set (4 environments) and tested on the remaining 20\%. PT-Net performance is comparable to SOTA.}
    \label{tab:sotaCompare}
    \vspace{-5pt}
\end{table*}

\subsection{Pedestrian Behavior Histograms}
Environment usage and social behavior characterization can also be done by observing the distribution of behaviors in each space. Figure \ref{fig:histograms} shows the pedestrian behavior histograms for ETH, ETH Hotel, UCY Zara1, and UCY Zara2 (along the column dimension) for $N=2,3$ people (along the row dimension). The histograms for ETH are significantly different from those of the other environments due to the difference in space usage (entryway to a building vs regular sidewalks). Pedestrians in ETH mainly exhibit horizontal leader-follower and walking side-by-side behaviors (N2:C15,23 and N3:C4,7) that allow them to enter or exit the building, while the other environments allow for more diversified movement. 

ETH Hotel, UCY Zara1, and UCY Zara 2 have higher concentrations of pedestrians exhibiting the group congregating (N2:C6,7 and N3:C13,24), walking side-by-side upwards (N2:C28 and N3:C25), and walking leader-follower or side-by-side diagonally downward (N2:C27 and N3:C16) behaviors than ETH. The spatial structure of ETH Hotel predisposes it towards side-by-side vertical behavior, over that of ETH, because it has a train stop at the top of the frame. However, UCY Zara1 and UCY Zara2 are more primarily dominated by horizontal leader-follower or side-by-side behaviors, as seen in ETH, because the side walk in front of the building is a more popular avenue than the alley at the edge of the frame. Even with this similarity, UCY Zara1 and UCY Zara2 have much more pedestrian congregation than ETH, showing that pedestrians in the ETH environment are more purposed in their movement or that loitering is not accepted in the space. Additionally, between UCY Zara1 and UCY Zara2, which take place in the same environment at different times, there is variation in pedestrian quantity showing a preference amongst pedestrians for which time of day they prefer to be more active in the space.

\subsection{Pedestrian Trajectory Prediction}

% \textbf{include experiment where we train with ABC test on D to see if we can learn how people behave generally from this, instead of just how people behave in a specific space; discuss results honestly in conclusion; useful to characterize a specific space, but also useful to have more general behavior analysis; train embedding on ABC but do full traj pred for D?; or train full on ABC then test on D for traj pred sight unseen?; to test generic-ness to new scenarios; may not be literally absolutely generic, but definitely at least generic to similar environments}
% \textbf{ are other people doing this????? mainly kothari2021interpretable?}

% \begin{figure}
%     \centering
%     \includegraphics[height=.49\textwidth, angle=-90]{ETH_1.png}
%     \includegraphics[height=.49\textwidth, angle=-90]{ETH_2.png}
%     \caption{PT-net future pedestrian trajectory predictions are shown in orange above superimposed on the corresponding video frame from the beginning of the sequence. The input sequence is shown in blue, and the ground truth prediction is shown in green. Notice that the directionality of the pedestrian is not lost. They are flowing from the blue (the input to the network) to the green (the ground truth for the predictions). \textbf{I feel like this figure doesn't really add anything... and the pedestrians are actually really hard to see anyway} }
%     \label{fig:framePred}
% \end{figure}

% \textbf{I should find the FDE also. and FDE would just be the loss of the last location?}% should I use those as the loss signals instead? would that be better? does it matter?}

We show the utility of our method on pedestrian trajectory prediction. Table \ref{tab:sotaCompare} shows a comparison of the metrics for PT-net and SOTA methods, where we show comparable performance on the ETH\cite{ETH} and UCY\cite{UCY} datasets.  For the SOTA methods, each network is trained with the leave one out strategy and tested on the remaining dataset. PT-net learns scene-specific features, so it is trained on 80\% of the total trajectory set (4 environments) and tested on the remaining 20\%. We provide metrics for the average displacement error (ADE) and final displacement error (FDE) of our method. For SOTA, we use officially reported results for ADE$_{20}$ and FDE$_{20}$, where the best of the k=20 sampled trajectories is chosen. %We also show that our method has significantly fewer parameters than either SOTA work. 
Figure \ref{fig:trajPerCluster} shows example trajectory predictions from a selection of social behavior clusters from the pedestrian behavior dictionary. The input trajectories are in blue, the ground truth future trajectories are in green, and the predictions are in orange. We predict plausible future trajectories for a myriad of socially complex scenarios using simple MLP networks. %Figure \ref{fig:framePred} shows two of these predictions on the corresponding frames from the ETH dataset. Figure \ref{fig:trajCompare} shows a qualitative comparison of our method with both D-LSTM\cite{kothari2021human} and Social-LSTM\cite{alahi2016social}. %\textbf{say more about how ours is better in the example when you get the figure} 
\section{Conclusion}
\vspace{-5pt}
In this paper, we proposed PT-net, a lightweight, unsupervised method for learning an interpretable pedestrian behavior dictionary for a given environment through trajectory clustering. Unsupervised methods remove the need for costly dataset labeling while allowing for the discovery of a dictionary containing diverse behavior patterns. With this dictionary, it is possible to characterize space usage and social behavior patterns to answer key questions in social science fields, like psychology and urban planning, using behavior maps and histograms to visualize the distribution of behaviors. We also demonstrate comparable performance to SOTA in trajectory prediction on the ETH and UCY datasets with a much simpler network. Decreasing the complexity and size of trajectory prediction methods is important for mobile computing and applications with limited computational resources.

\section*{Acknowledgements}
This work was supported by grant NSF NRT NRT-FW-HTF: Socially Cognizant Robotics for a Technology Enhanced Society (SOCRATES), No. 2021628.

% \noindent This work was done under the SOCRATES NSF NRT grant (grant number here!!!) and with the help of Jacob Feldman and Clint Andrews. 

% \begin{itemize}
%     \item unsupervised representation learning that's clustering based is new and exciting and necessary trend
%     \item lightweight implementations are better for mobile/limited computational power devices
%     \item traj pred important for autonomous robots and needs to be lightweight
%     \item characterizing a particular space has high value and allows for real applications in urban systems(?)
% \end{itemize}

% \clearpage

%%%%%%%%% REFERENCES
{\small
\bibliographystyle{ieee_fullname}
\bibliography{egbib}
}

\end{document}